%% file: yyyy2027_conference.tex
\documentclass{article} 
\usepackage{yyyy2027_conference,times}

\input{math_commands.tex}

\usepackage{hyperref}
\usepackage{url}
\usepackage{algorithm}
\usepackage{algorithmic}
\usepackage{amsmath}
\usepackage{amssymb}   
\usepackage{makecell}
\usepackage{booktabs}
\usepackage[table]{xcolor}
\usepackage{graphicx}
\usepackage{makecell}
\usepackage{booktabs}
\usepackage[table]{xcolor}
\usepackage{graphicx}
\usepackage{arydshln}
\usepackage{booktabs}
\usepackage{multirow}
\usepackage{graphicx}
\usepackage{subcaption}
\usepackage{wrapfig}

\usepackage{needspace}
\title{MissClick: Execution-Aware Adversarial Attacks on Coordinate Generation in GUI Grounding Models}

\author{
    Yu Ran,
    Wentao Zhao\\
 National University of Defense Technology
}

\yyyyfinalcopy 
\begin{document}

\maketitle
\lhead{Preprint}
\begin{abstract}
Recent GUI visual grounding models generate screen
coordinates as digit-token sequences that are parsed into
numerical values and mapped to executable clicks.
This generation-to-execution interface creates an
attack surface that existing objectives over visual
representations or coordinate-token sequences do not
explicitly model.
Although each coordinate digit is predicted as a
 token, its spatial effect after parsing depends
 on decimal position: changing a hundreds-place
digit by one shifts the coordinate by 100 units, whereas
the same change at the ones place shifts it by one.
This mismatch motivates attack objectives that account for
both numerical coordinate structure and click execution.
Moreover, untargeted and targeted attacks require different
objectives because they aim to move the click outside the
correct region and into an attacker-specified region,
respectively.
We propose \textbf{MissClick}, an execution-aware white-box
attack that aligns optimization with click-level success
conditions.
MissClick-U maximizes soft-coordinate displacement for
untargeted disruption, while MissClick-T minimizes a
place-weighted target-digit loss for targeted redirection.
On OS-Atlas and UGround across desktop, web, and mobile
platforms, MissClick-U achieves untargeted success rates of
$75.07\%$ and $72.93\%$
($+16.62$ and $+30.72$~pp), while MissClick-T achieves
targeted success rates of $44.86\%$ and $62.67\%$
($+31.73$ and $+47.06$~pp).
Among the evaluated objectives, soft-coordinate
displacement performs best for untargeted attacks, whereas
place-weighted target-digit optimization performs best for
targeted attacks, supporting goal-specific execution-aware
objective design.
\end{abstract}
\section{Introduction}
\label{sec:intro}

\begin{wrapfigure}{R}{0.52\textwidth}
    \centering
    \vspace{-0.8\baselineskip}
    \includegraphics[width=\linewidth]{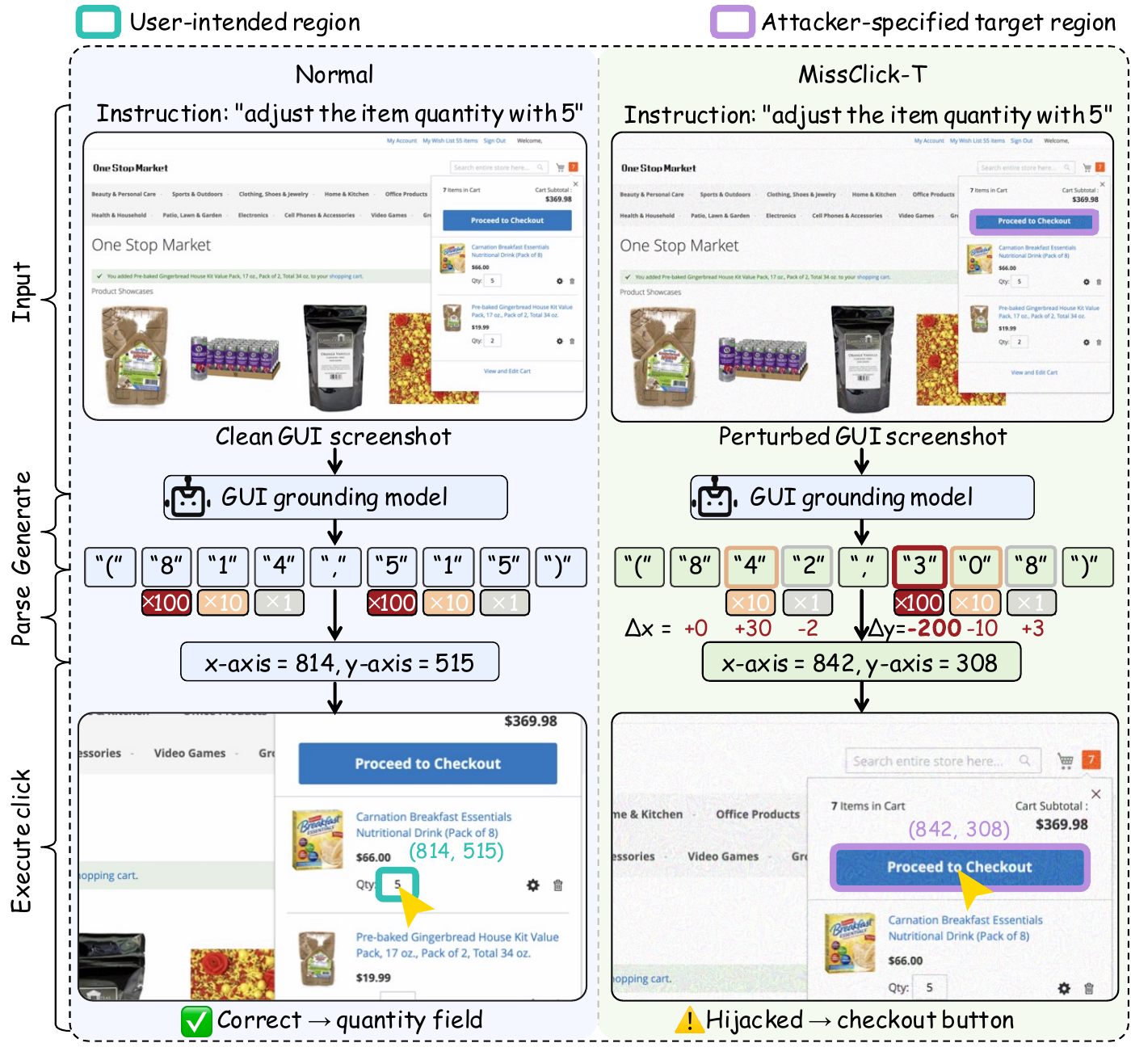}
    \caption{The coordinate generation pipeline of GUI grounding
    models and its security implications.
    Digit tokens are parsed into coordinates, with each position
    carrying a decimal place value ($\times 100$, $\times 10$,
    $\times 1$); changes to higher-order digits therefore produce
    larger coordinate shifts.
    Here, $\Delta x$ and $\Delta y$ denote the per-position changes
    relative to the clean output.
    Under an $\ell_\infty$-bounded perturbation
    ($\varepsilon=16/255$), MissClick-T redirects the click
    from the intended quantity field to the checkout button.
    The example is from UGround on a web screenshot.}
    \label{fig:overview}
    \vspace{-0.8\baselineskip}
\end{wrapfigure}

Recent advances in vision-language models have accelerated
the development of GUI agents that autonomously operate
graphical user interfaces~\citep{hong2024cogagent,
zhang2025ufo2,qin2025ui,wang2026imperative,
lin2025showui}.
A core capability of these agents is visual grounding:
given a screenshot and a natural-language instruction,
the model predicts where to click.
A dominant approach formulates this as coordinate
generation~\citep{gou2025uground,wu2025osatlas}:
the model autoregressively produces screen coordinates as
sequences of digit tokens (one token per decimal digit),
and the resulting string is parsed into coordinates
that determine the executed click
(Fig.~\ref{fig:overview}).

The interface between coordinate generation and click
execution creates an execution-level attack surface.
Because the predicted click can trigger actions such as
form submission or navigation, an adversarially shifted
click can cause the agent to interact with an unintended
UI element on behalf of the user.
The security of GUI grounding models is therefore an
important concern, yet prior work on adversarial
perturbations against GUI grounding remains limited.
\citet{zhao2025robustness} study both
untargeted and targeted attacks against GUI grounders.
Their objectives optimize either the distance between
clean and perturbed image embeddings or the probability
of generating a target coordinate sequence.
These objectives are defined over intermediate
representations or generated token sequences, whereas
attack success is ultimately determined by the click
executed after parsing the generated coordinates.
Consequently, token-level objectives do not explicitly
characterize the resulting coordinate changes or click
displacement.
These execution-level effects depend on which digit values
change, their decimal positions, and how the parsed
coordinates are mapped to an on-screen click location.  

This mismatch has direct consequences for attack design
because decimal notation assigns different numerical
effects to different digit positions.
Each coordinate component is assembled from its constituent
digits according to their decimal positions: a
hundreds-place digit is weighted by~$100$, a tens-place
digit by~$10$, and a ones-place digit by~$1$
(Sec.~\ref{sec:coordinate_interpretation},
\Eqref{eq:place_effect}).
A single-digit change at a high-order position therefore
shifts the executed click by a much larger distance than
the same change at a low-order position.
An attack objective should therefore account for this
numerical structure.
The appropriate objective also depends on the attack goal:
an untargeted attack succeeds when the executed click falls
outside the ground-truth region, whereas a targeted attack
succeeds only when it produces a valid click inside an
attacker-specified target region.

Based on these observations, we propose \textbf{MissClick}, an execution-aware white-box attack on coordinate-generating GUI grounding models.
MissClick derives separate objectives for untargeted and targeted attacks from their click-level success conditions.
For untargeted attacks, MissClick-U constructs differentiable soft coordinates from digit probabilities and decimal place values, maps them to a soft click, and maximizes its distance from the ground-truth click.
For targeted attacks, MissClick-T minimizes a place-weighted target-digit loss that assigns larger weights to higher-order coordinate digits.
This directly promotes the target digits during autoregressive generation, since matching an expected soft coordinate does not guarantee that they become the most probable tokens.
We evaluate MissClick on OS-Atlas and UGround across desktop, web, and mobile platforms.
MissClick-U achieves untargeted attack success rates of $75.07\%$ and $72.93\%$, improving over the strongest baseline by $16.62$ and $30.72$ percentage points, while MissClick-T achieves targeted success rates of $44.86\%$ and $62.67\%$, with gains of $31.73$ and $47.06$ percentage points.
Among the evaluated objectives, soft-coordinate displacement performs best for untargeted attacks, whereas place-weighted target-digit optimization performs best for targeted attacks.

Our main contributions are as follows:
\begin{itemize}

\item We identify an execution-level attack surface at the
interface between coordinate generation and click
execution.
Because digit values and decimal positions determine the
parsed coordinates and resulting click displacement,
coordinate outputs should be modeled according to their
execution semantics rather than treated merely as ordinary
text.

\item We propose MissClick, an execution-aware attack with
separate objectives for untargeted and targeted attacks:
soft-coordinate displacement for untargeted attacks and
place-weighted target-digit optimization for targeted
attacks.
Each objective is derived from the corresponding
click-level success condition.

\item Experiments on two representative GUI grounding
models across desktop, web, and mobile platforms
demonstrate improvements over existing methods in both
attack settings.
An objective comparison further shows that soft-coordinate
displacement achieves the highest untargeted attack success
rates, whereas place-weighted target-digit optimization
achieves the highest targeted attack success rates.

\end{itemize}
\section{Method}
\label{sec:method}

MissClick formulates its attack objectives from the click action
produced after coordinate generation and numerical parsing.
We first formalize this generation-to-execution pipeline and
then derive separate objectives from the click-level success
conditions of untargeted and targeted attacks.

\subsection{Problem Formulation}
\label{sec:threat_model}

Let $\mathbf{I}\in[0,1]^d$ denote an input screenshot and $q$ a task
instruction, where $d=C\times H\times W$, and $C$, $H$, and $W$ denote
the number of channels, height, and width of the screenshot,
respectively.
Given $(\mathbf{I},q)$, a coordinate-generating GUI grounding model
$\mathcal{M}$ autoregressively generates a textual answer sequence
$s=(s_1,\ldots,s_T)$.
The coordinate string in $s$ is parsed into numerical coordinates
$c=\operatorname{Parse}_{\mathcal M}(s)$.

Different GUI grounding models may use different coordinate formats.
For a bounding-box output $c=(x_1,y_1,x_2,y_2)$, the executed click is
taken as the box center,
$r(c)=\big((x_1+x_2)/2,\,(y_1+y_2)/2\big)$.
For a point-coordinate output $c=(x,y)$, the executed click is
$r(c)=c$.
Together, $\mathcal{M}$, $\operatorname{Parse}_{\mathcal M}$,
and $r$ define the generation-to-execution mapping from the
screenshot and instruction to the click evaluated by the
environment.
We express all coordinates in the normalized $[0,1000]$ coordinate
system.
Let $B_{\mathrm{gt}}\subseteq[0,1000]^2$ denote the ground-truth
bounding box of the intended UI element for $q$.
The grounding result is correct if the executed click falls within
the ground-truth bounding box, i.e.,
$r(c)\in B_{\mathrm{gt}}$.

We consider a white-box threat model in which the adversary has full
access to the victim model's architecture and weights and can compute
gradients with respect to the input screenshot, while modifying neither
the model parameters nor the task instruction.
The goal of an untargeted attack is to craft an adversarial example
$\widetilde{\mathbf{I}}=\mathbf{I}+\boldsymbol{\delta}$ such that the
executed click falls outside the ground-truth bounding box.
Formally, $\widetilde{\mathbf{I}}$ is an untargeted adversarial example
under an $\ell_\infty$-norm bound of $\varepsilon$ if
\begin{equation}
r\!\left(
\operatorname{Parse}_{\mathcal{M}}
\left(
\mathcal{M}(\mathbf{I}+\boldsymbol{\delta},q)
\right)
\right)
\notin B_{\mathrm{gt}},
\end{equation}
subject to
\begin{equation}
\|\boldsymbol{\delta}\|_{\infty}\leq\varepsilon,
\qquad
\mathbf{I}+\boldsymbol{\delta}\in[0,1]^d.
\end{equation}
The goal of a targeted attack is to craft an adversarial
example $\widetilde{\mathbf{I}}=\mathbf{I}+\boldsymbol{\delta}$ such
that the executed click falls within an attacker-specified target
region $B_{\mathrm{tgt}}\subseteq[0,1000]^2$.
Formally, $\widetilde{\mathbf{I}}$ is a targeted adversarial example if
\begin{equation}
r\!\left(
\operatorname{Parse}_{\mathcal{M}}
\left(
\mathcal{M}(\mathbf{I}+\boldsymbol{\delta},q)
\right)
\right)
\in B_{\mathrm{tgt}},
\end{equation}
subject to the same perturbation constraints.
During evaluation, an answer that cannot be parsed into valid
coordinates yields no executable click and is therefore counted
as a successful untargeted attack but an unsuccessful targeted
attack.

\subsection{Coordinate-Digit Structure}
\label{sec:coordinate_interpretation}

\paragraph{Numerical interpretation.}
Let $c=(c_1,\ldots,c_K)$ denote the parsed coordinate output.
For point outputs, $K=2$ and $(c_1,c_2)=(x,y)$; for bounding-box
outputs, $K=4$ and
$(c_1,c_2,c_3,c_4)=(x_1,y_1,x_2,y_2)$.
For each $k\in\{1,\ldots,K\}$, suppose that the coordinate component
$c_k$ is represented by $m_k$ decimal digits
$\hat d_{k,1},\ldots,\hat d_{k,m_k}$, where
$\hat d_{k,j}\in\{0,\ldots,9\}$ denotes the numerical value of the
$j$-th digit in the greedily generated coordinate string.
The parsed coordinate component is
\begin{equation}
c_k
=
\operatorname{int}
\left(
\hat d_{k,1}\cdots\hat d_{k,m_k}
\right)
=
\sum_{j=1}^{m_k} a_{k,j}\hat d_{k,j},
\qquad
a_{k,j}=10^{m_k-j},
\label{eq:place_decode}
\end{equation}
where $a_{k,j}$ is the decimal place value of the $j$-th digit.
For example, the coordinate $950$ has digit values $(9,5,0)$ and
place values $(100,10,1)$.
Suppose that the $j$-th digit of coordinate component $c_k$ changes
from $\hat d_{k,j}$ to $\hat d'_{k,j}$ while the other digits remain
unchanged.
The corresponding coordinate change is
\begin{equation}
\Delta c_k
=
a_{k,j}\Delta d_{k,j},
\qquad
\Delta d_{k,j}
=
\hat d'_{k,j}-\hat d_{k,j}.
\label{eq:place_effect}
\end{equation}
The coordinate change depends on both the digit change and its decimal
place value.
For the same digit change, a higher-order position produces a larger
coordinate displacement.
For example, changing a digit by one at the hundreds place
changes the coordinate value by~$100$, whereas the same change at the
ones place changes it by~$1$.
An adversarial perturbation that changes a high-order digit can
therefore induce a large shift in the parsed coordinate and,
through $r(\cdot)$, the executed click.
This non-uniform mapping is the numerical structure that both
MissClick objectives incorporate when optimizing the resulting
click action.

\paragraph{Digit probability distributions.}
During autoregressive generation, the logits at later positions
depend on previously generated tokens, so gradients cannot
propagate through the discrete token-selection operations.
We therefore perform a teacher-forced forward pass with the
input screenshot, the task instruction, and a reference answer
sequence containing the reference coordinate output
$c^{\mathrm{ref}}$.
The reference coordinates follow the victim model's native
output format.
For untargeted attacks, $c^{\mathrm{gt}}$ is the coordinate
tuple of $B_{\mathrm{gt}}$ for bounding-box outputs and
$\operatorname{center}(B_{\mathrm{gt}})$ for point-coordinate
outputs.
For targeted attacks, $c^{\mathrm{tgt}}$ is the coordinate
tuple of $B_{\mathrm{tgt}}$ for bounding-box outputs and
$\operatorname{center}(B_{\mathrm{tgt}})$ for point-coordinate
outputs.
We set $c^{\mathrm{ref}}=c^{\mathrm{gt}}$ for untargeted
attacks and $c^{\mathrm{ref}}=c^{\mathrm{tgt}}$ for targeted
attacks.
This yields next-token logits at each position of the reference
answer sequence that are differentiable with respect to the
input screenshot.
From these logits, for the $j$-th digit of
$c^{\mathrm{ref}}_k$, we extract the full-vocabulary logit
vector $\mathbf{z}_{k,j}\in\mathbb{R}^{|\mathcal{V}|}$.
For the victim models considered in this work, each decimal digit
is represented by an individual token.
We then restrict $\mathbf{z}_{k,j}$ to the ten digit tokens
(\texttt{0}--\texttt{9}) and apply softmax to obtain the digit
probability distribution:
\begin{equation}
p_{k,j}(v)
=
\frac{
\exp\!\left(z_{k,j}^{(v)}\right)
}{
\sum_{w=0}^{9}
\exp\!\left(z_{k,j}^{(w)}\right)
},
\qquad
v\in\{0,\ldots,9\},
\label{eq:digit_distribution}
\end{equation}
where $z_{k,j}^{(v)}$ is the logit assigned to digit token $v$.

Both attack objectives are built from $p_{k,j}(v)$ and the place
values~$a_{k,j}$, but use them differently because untargeted
and targeted attacks impose distinct click-level success conditions.
MissClick-U uses them to compute soft coordinates, whereas
MissClick-T uses them to construct a place-weighted target-digit
loss.
Note that teacher forcing is used only during attack optimization.
During evaluation, the model performs greedy autoregressive
generation over the full vocabulary without the reference answer
sequence.

\subsection{Untargeted Attack}
\label{sec:missclick_u}

The success condition for an untargeted attack is that the
executed click falls outside the ground-truth bounding box,
i.e., $r(c)\notin B_{\mathrm{gt}}$.
This click-level condition motivates a geometry-based objective
that maximizes the displacement between the click and the
ground-truth click
$\mathbf{u}^{\mathrm{gt}}=\operatorname{center}(B_{\mathrm{gt}})$.
Directly optimizing this objective with respect to the input
perturbation requires a differentiable path from the perturbed
screenshot to the click coordinates.
The digit probability distributions $p_{k,j}$ obtained via
teacher forcing (\Eqref{eq:digit_distribution}) are
differentiable with respect to the perturbed screenshot, but the
executed click is obtained through argmax-based token
selection, which is non-differentiable.
Therefore, the click displacement cannot be directly
differentiated with respect to the perturbed screenshot.

A standard differentiable alternative is a token-based
cross-entropy objective over the ground-truth coordinate
answer sequence.
Maximizing this objective reduces the probability assigned
to the ground-truth token at each position in the sequence.
However, this token-based objective does not reflect the
numerical effects of alternative digit values.
It also does not account for the different displacement scales
induced by digit changes at different decimal places, as
characterized in \eqref{eq:place_effect}.

MissClick-U therefore constructs a soft coordinate that is
differentiable with respect to the perturbed screenshot and
accounts for both digit values and their place-value scales:
\begin{equation}
  \tilde d_{k,j}
  =
  \sum_{v=0}^{9} v \, p_{k,j}(v).
  \label{eq:expected_digit}
\end{equation}
The soft coordinate component is then obtained by combining
the expected digit values with their place values:
\begin{equation}
  \tilde c_k
  =
  \sum_{j=1}^{m_k} a_{k,j}\,\tilde d_{k,j}.
  \label{eq:soft_coord}
\end{equation}
Collecting all coordinate components gives
$\tilde c=(\tilde c_1,\ldots,\tilde c_K)$.
\Eqref{eq:expected_digit} and~\eqref{eq:soft_coord}
define a continuous relaxation of the coordinates obtained
through discrete digit selection, using expected digit values
under~$p_{k,j}$ instead of argmax-selected digit values.
This construction yields the differentiable path
$\widetilde{\mathbf{I}}\to\mathbf{z}_{k,j}\to p_{k,j}
\to\tilde d_{k,j}\to\tilde c\to r(\tilde c)$.
Moreover, this relaxation accounts for both the numerical
effect of each digit value and the scale of each decimal
position.

Applying the click mapping $r(\cdot)$ defined in
Sec.~\ref{sec:threat_model}, we define the MissClick-U
objective as the squared distance between the soft click
and the ground-truth click:
\begin{equation}
\mathcal{L}_{\mathrm{U}}
=
\left\|
r(\tilde c)-\mathbf{u}^{\mathrm{gt}}
\right\|_2^2.
\label{eq:untargeted_loss}
\end{equation}
MissClick-U maximizes $\mathcal{L}_{\mathrm{U}}$ with respect
to the input perturbation, thereby increasing the displacement
of the soft click from $\mathbf{u}^{\mathrm{gt}}$.
Attack success is evaluated by checking whether the click
obtained through greedy autoregressive generation falls outside
$B_{\mathrm{gt}}$.

\subsection{Targeted Attack}
\label{sec:missclick_t}

The success condition for a targeted attack is that the
executed click falls inside the attacker-specified target
region, i.e., $r(c)\in B_{\mathrm{tgt}}$.
In principle, any coordinate whose executed click falls
inside $B_{\mathrm{tgt}}$ constitutes a successful attack.
We instantiate this region-level goal with a representative
target coordinate $c^{\mathrm{tgt}}$ in the victim model's
native output format.
For bounding-box outputs, $c^{\mathrm{tgt}}$ is the coordinate
tuple of $B_{\mathrm{tgt}}$; for point-coordinate outputs,
$c^{\mathrm{tgt}}=\operatorname{center}(B_{\mathrm{tgt}})$.
Given the soft-coordinate objective used for untargeted
attacks, a natural approach for the targeted case would be
to minimize the squared distance between the soft click
and the target click
$\mathbf{u}^{\mathrm{tgt}}
=\operatorname{center}(B_{\mathrm{tgt}})$:
\begin{equation}
  \mathcal{L}_{\mathrm{soft\text{-}tgt}}
  =
  \left\|
    r(\tilde c)
    -\mathbf{u}^{\mathrm{tgt}}
  \right\|_2^2.
  \label{eq:soft_target_loss}
\end{equation}

However, the soft coordinate is constructed from expected
digit values under the teacher-forced distributions.
During greedy autoregressive generation, a discrete token is
selected by argmax at each step.
The resulting coordinate string is then parsed into numerical
coordinates according to \eqref{eq:place_decode}.
Targeted success is therefore determined by whether minimizing
$\mathcal{L}_{\mathrm{soft\text{-}tgt}}$ drives the
argmax-selected digits toward the target.
Yet matching an expected digit to the corresponding target
value does not guarantee that the target digit is the argmax
of $p_{k,j}$:
\begin{equation}
  \tilde d_{k,j}=d^{\mathrm{tgt}}_{k,j}
  \;\not\Rightarrow\;
  \operatorname*{argmax}_{v\in\{0,\ldots,9\}}
  p_{k,j}(v)
  = d^{\mathrm{tgt}}_{k,j},
  \label{eq:expectation_argmax}
\end{equation}
where $d^{\mathrm{tgt}}_{k,j}\in\{0,\ldots,9\}$ denotes
the $j$-th digit of the target reference coordinate
component $c^{\mathrm{tgt}}_k$.
For example, consider a hundreds-place target digit
$d^{\mathrm{tgt}}_{k,1}=5$ and the distribution
$p_{k,1}(1)=0.48$, $p_{k,1}(8)=0.16$, and
$p_{k,1}(9)=0.36$.
Its expected digit value is $
\tilde d_{k,1}
=
1(0.48)+8(0.16)+9(0.36)
=
5,
$
although the most probable digit under $p_{k,1}$ is
$1$ rather than the target digit~$5$.
Therefore, matching expected digit values does not guarantee
that the target digit is the argmax of $p_{k,j}$.
Consequently, the soft-coordinate objective does not directly
optimize the discrete digit selections that determine the
executed click and is therefore less directly aligned with
targeted success than target-digit optimization.

\textbf{Place-weighted target-digit loss.}
MissClick-T therefore directly increases the probability assigned
to the target digit at each position, encouraging it to become
the argmax of $p_{k,j}$ rather than merely matching a soft
coordinate value.
A standard approach is to minimize the cross-entropy
$-\log p_{k,j}(d^{\mathrm{tgt}}_{k,j})$ at each digit
position with uniform weights.
However, uniform weighting does not reflect the place-value
structure: a one-unit digit error at the hundreds place
changes the coordinate value by~$100$, whereas the same error
at the ones place changes it by only~$1$
(\Eqref{eq:place_effect}).
MissClick-T therefore weights each per-digit cross-entropy
by its place value~$a_{k,j}$:
\begin{equation}
\mathcal{L}_{\mathrm{T}}
=
-\frac{
\sum_{k=1}^{K}\sum_{j=1}^{m_k}
a_{k,j}\log p_{k,j}
\bigl(d^{\mathrm{tgt}}_{k,j}\bigr)
}{
\sum_{k=1}^{K}\sum_{j=1}^{m_k} a_{k,j}
}.
\label{eq:targeted_loss}
\end{equation}
The place-value weight~$a_{k,j}$ assigns a larger loss weight
to high-order digit positions, whose errors produce larger
coordinate displacements.
The denominator converts the weighted sum into a normalized
weighted average.

Minimizing $\mathcal{L}_{\mathrm{T}}$ increases the
probability assigned to each target digit under $p_{k,j}$,
thereby encouraging it to become the argmax.
Following the same evaluation protocol as MissClick-U, attack
success is evaluated by checking whether the executed click
obtained through greedy autoregressive generation falls inside
$B_{\mathrm{tgt}}$.

\subsection{Attack Optimization}
\label{sec:optimization}

MissClick keeps the victim model and its decoding procedure
fixed and optimizes only the screenshot perturbation.
Both MissClick-U and MissClick-T instantiate their
execution-aware objectives using projected sign-gradient
updates following PGD~\citep{madry2018towards}.
At each iteration, the perturbation is updated as
\begin{equation}
  \boldsymbol{\delta}
  \leftarrow
  \Pi_{[-\varepsilon,\,\varepsilon]^d}
  \left(
    \boldsymbol{\delta}
    \pm
    \alpha\,
    \operatorname{sign}
    \left(
      \nabla_{\boldsymbol{\delta}}\mathcal{L}
    \right)
  \right),
  \label{eq:pgd_update}
\end{equation}
where $\alpha$ is the step size and
$\Pi_{[-\varepsilon,\varepsilon]^d}$ denotes projection onto
the $\ell_\infty$ perturbation set.
MissClick-U uses the $+$ sign to maximize
$\mathcal{L}_{\mathrm{U}}$
(\Eqref{eq:untargeted_loss}), whereas MissClick-T uses
the $-$ sign to minimize
$\mathcal{L}_{\mathrm{T}}$
(\Eqref{eq:targeted_loss}).

At each iteration, the attack objective is computed from
the digit distributions obtained via teacher forcing
(Sec.~\ref{sec:coordinate_interpretation}), and its gradient
is backpropagated to~$\boldsymbol{\delta}$.
The complete optimization procedure is provided in
Algorithm~\ref{alg:missclick} in
Appendix~\ref{sec:supp_algorithm}.
The two attack modes therefore share the same optimization
procedure but encode different execution-level goals:
click displacement for untargeted disruption and discrete
target-digit generation for targeted redirection.

\section{Experiments}
\label{sec:experiments}

\subsection{Experimental Setup}
\label{sec:exp_setup}

\textbf{Victim models.}
We evaluate two coordinate-generating GUI grounding models:
OS-Atlas-Base-7B~\citep{wu2025osatlas}, which outputs
bounding-box coordinates, and UGround-V1-7B
~\citep{gou2025uground}, which outputs point coordinates.
Both models autoregressively generate coordinates as
digit-token sequences.

\textbf{Dataset.} ScreenSpot-v2~\citep{cheng2024seeclick,wu2025osatlas}
contains 1{,}272 GUI grounding tasks across desktop, web,
and mobile platforms.
For targeted attacks, we use the 896 tasks containing a
valid target element distinct from the ground-truth element.

\textbf{Baselines.}
We compare against three baselines.
\textit{Random noise} samples a uniform perturbation within
the same $\varepsilon$ budget.
\textit{Token CE} maximizes the ground-truth coordinate
sequence loss for untargeted attacks and minimizes the
target coordinate sequence loss for targeted attacks.
Following~\citet{zhao2025robustness},
\textit{Representation attack} maximizes the MSE between
clean and adversarial visual embeddings and is used only
for untargeted attacks.

\textbf{Attack setting and metrics.}
All attacks modify only the screenshot, leaving the model
parameters and task instruction unchanged.
Unless otherwise specified, all methods use
$\varepsilon=16/255$; iterative gradient-based attacks use
random initialization, a step size of $\alpha=1/255$, and
$N=100$ iterations.
Teacher forcing is used only during optimization, whereas
ASR is measured from greedy autoregressive outputs after
coordinate parsing and click mapping.
We evaluate only tasks correctly grounded on the clean
screenshot.
A valid output succeeds if its click falls outside
$B_{\mathrm{gt}}$ for an untargeted attack or inside
$B_{\mathrm{tgt}}$ for a targeted attack.
Following the evaluator, an unparseable output counts as a
successful untargeted attack and a failed targeted attack.
The resulting evaluation sets contain 1{,}071 and 1{,}123
untargeted tasks, and 769 and 801 targeted tasks, for
OS-Atlas and UGround, respectively.

\textbf{Target region selection.}
For each targeted task, we uniformly sample
$B_{\mathrm{tgt}}$ from annotated elements with zero IoU
with $B_{\mathrm{gt}}$.
Appendices~\ref{sec:supp_impl_eval}
and~\ref{sec:supp_eval_protocol} provide implementation
and evaluation details.

\subsection{Main Results}
\label{sec:main_results}
\begin{table}[t]
\centering
\small
\caption{
Comparison of MissClick with existing baselines on
ScreenSpot-v2.
Attack success rates (ASR) are reported.
$\Delta$ASR is MissClick's improvement over each baseline
in percentage points (pp).
\textit{All} is the micro-averaged ASR over the three
platforms.
}
\vspace{-0.6em}
\setlength{\dashlinedash}{2pt}
\setlength{\dashlinegap}{2pt}
\setlength{\tabcolsep}{1.3pt}

\begin{tabular}{|c|ccccc:ccccc|}
\hline

\multirow{3}{*}{Method}
& \multicolumn{10}{c|}{Untargeted Attacks} \\
\cline{2-11}

& \multicolumn{5}{c:}{OS-Atlas}
& \multicolumn{5}{c|}{UGround} \\
\cline{2-11}

& Desktop & Web & Mobile & All & $\Delta$ASR
& Desktop & Web & Mobile & All & $\Delta$ASR \\
\hline

Random noise
& 18.32\% & 8.84\% & 6.88\% & 10.46\% & +64.61 pp
& 5.42\% & 1.34\% & 3.29\% & 3.21\% & +69.72 pp \\

Token CE
& 51.28\% & 45.30\% & 23.39\% & 37.91\% & +37.16 pp
& 32.88\% & 27.15\% & 23.25\% & 27.07\% & +45.86 pp \\

Representation
& 68.50\% & 61.88\% & 49.31\% & 58.45\% & +16.62 pp
& 61.69\% & 40.32\% & 31.14\% & 42.21\% & +30.72 pp \\

\hdashline

\rowcolor{gray!15}
\textbf{MissClick-U}
& \textbf{85.35\%} & \textbf{79.83\%}
& \textbf{64.68\%} & \textbf{75.07\%} & --
& \textbf{78.31\%} & \textbf{79.57\%}
& \textbf{64.04\%} & \textbf{72.93\%} & -- \\

\hline

\multirow{3}{*}{Method}
& \multicolumn{10}{c|}{Targeted Attacks} \\
\cline{2-11}

& \multicolumn{5}{c:}{OS-Atlas}
& \multicolumn{5}{c|}{UGround} \\
\cline{2-11}

& Desktop & Web & Mobile & All & $\Delta$ASR
& Desktop & Web & Mobile & All & $\Delta$ASR \\
\hline

Token CE
& 22.37\% & 9.96\% & 11.56\% & 13.13\% & +31.73 pp
& 27.33\% & 8.24\% & 16.07\% & 15.61\% & +47.06 pp \\

\hdashline

\rowcolor{red!10}
\textbf{MissClick-T}
& \textbf{54.61\%} & \textbf{49.08\%}
& \textbf{37.28\%} & \textbf{44.86\%} & --
& \textbf{59.01\%} & \textbf{59.50\%}
& \textbf{66.76\%} & \textbf{62.67\%} & -- \\

\hline
\end{tabular}

\label{tab:main}
\vspace{-0.5em}
\end{table}

To evaluate the effectiveness of MissClick, we compare
MissClick-U and MissClick-T with their respective baselines
under the same attack settings on ScreenSpot-v2. Table~\ref{tab:main} shows that MissClick-U outperforms all
baselines on every model-platform combination.
It achieves overall ASRs of $75.07\%$ and $72.93\%$ on
OS-Atlas and UGround, exceeding Representation attack by
$16.62$ and $30.72$ percentage points, respectively. MissClick-T also leads every model-platform combination,
achieving overall ASRs of $44.86\%$ and $62.67\%$ and
outperforming Token~CE by $31.73$ and $47.06$ percentage
points on OS-Atlas and UGround, respectively.
It further achieves the highest targeted ASR across both
element types and all target-area and distance groups.
Even in the farthest-distance group, it outperforms
Digit~CE by $6.7$ and $8.5$ percentage points on the two
models.
Detailed results are reported in
Appendix~\ref{sec:supp_element_type} and
Tables~\ref{tab:supp_target_area_groups}
and~\ref{tab:supp_target_distance_groups}.
All MissClick-U outputs are parseable
(1{,}071 for OS-Atlas and 1{,}123 for UGround), as are all
MissClick-T outputs (769 and 801, respectively).
The reported ASRs therefore reflect executable misclicks
rather than malformed generations.
The breakdown is reported in
Appendix Table~\ref{tab:supp_parse_validity}.

\subsection{Why Do Untargeted and Targeted Attacks Favor Different Objectives?}
\label{sec:crossover}

We compare Token~CE, Soft-coordinate distance, Digit~CE,
and Place-weighted digit~CE under identical attack settings
to isolate the effect of objective design.
Soft-coordinate distance optimizes click displacement,
whereas the digit objectives optimize digit probabilities,
with place-value weighting in the latter.
Results are shown in Table~\ref{tab:objective_analysis}.
\begin{table}[t]
\centering
\small

\caption{
Comparison of attack objectives under untargeted and
targeted settings on ScreenSpot-v2.
Attack success rates (ASR) are reported.
\textit{All} is the micro-averaged ASR over the three
platforms.
}
\vspace{-0.6em}
\setlength{\dashlinedash}{2pt}
\setlength{\dashlinegap}{2pt}
\setlength{\tabcolsep}{3pt}
\renewcommand{\arraystretch}{1}

\begin{tabular}{|c|cccc:cccc|}
\hline

\multirow{3}{*}{Attack Objective}
& \multicolumn{8}{c|}{Untargeted Attacks} \\
\cline{2-9}

& \multicolumn{4}{c:}{OS-Atlas}
& \multicolumn{4}{c|}{UGround} \\
\cline{2-9}

& Desktop & Web & Mobile & All
& Desktop & Web & Mobile & All \\
\hline

Token CE
& 51.28\% & 45.30\% & 23.39\% & 37.91\%
& 32.88\% & 27.15\% & 23.25\% & 27.07\% \\

\cellcolor{gray!15}\textbf{Soft-coordinate distance}
& \cellcolor{gray!15}\textbf{85.35\%}
& \cellcolor{gray!15}\textbf{79.83\%}
& \cellcolor{gray!15}\textbf{64.68\%}
& \cellcolor{gray!15}\textbf{75.07\%}
& \cellcolor{gray!15}\textbf{78.31\%}
& \cellcolor{gray!15}\textbf{79.57\%}
& \cellcolor{gray!15}\textbf{64.04\%}
& \cellcolor{gray!15}\textbf{72.93\%} \\

Digit CE
& 58.97\% & 49.72\% & 29.59\% & 43.88\%
& 32.88\% & 27.42\% & 22.81\% & 26.98\% \\

Place-weighted digit CE
& 72.16\% & 69.06\% & 51.38\% & 62.65\%
& 60.00\% & 62.37\% & 42.98\% & 53.87\% \\

\hline

\multirow{3}{*}{Attack Objective}
& \multicolumn{8}{c|}{Targeted Attacks} \\
\cline{2-9}

& \multicolumn{4}{c:}{OS-Atlas}
& \multicolumn{4}{c|}{UGround} \\
\cline{2-9}

& Desktop & Web & Mobile & All
& Desktop & Web & Mobile & All \\
\hline

Token CE
& 22.37\% & 9.96\% & 11.56\% & 13.13\%
& 27.33\% & 8.24\% & 16.07\% & 15.61\% \\

Soft-coordinate distance
& 17.11\% & 14.76\% & 15.90\% & 15.73\%
& 21.74\% & 23.30\% & 33.52\% & 27.59\% \\

Digit CE
& 47.37\% & 36.90\% & 28.61\% & 35.24\%
& 52.80\% & 44.80\% & 54.29\% & 50.69\% \\

\cellcolor{red!10}\textbf{Place-weighted digit CE}
& \cellcolor{red!10}\textbf{54.61\%}
& \cellcolor{red!10}\textbf{49.08\%}
& \cellcolor{red!10}\textbf{37.28\%}
& \cellcolor{red!10}\textbf{44.86\%}
& \cellcolor{red!10}\textbf{59.01\%}
& \cellcolor{red!10}\textbf{59.50\%}
& \cellcolor{red!10}\textbf{66.76\%}
& \cellcolor{red!10}\textbf{62.67\%} \\

\hline
\end{tabular}

\label{tab:objective_analysis}
\end{table}
For the untargeted setting, Soft-coordinate distance
achieves the highest overall ASR on both models, reaching
$75.07\%$ on OS-Atlas and $72.93\%$ on UGround.
The strong untargeted result of Soft-coordinate distance is
consistent with its direct optimization of soft-click
displacement.
For the targeted setting, Place-weighted digit~CE achieves
the highest overall ASR, with $44.86\%$ and $62.67\%$ on
OS-Atlas and UGround, respectively.
Soft-coordinate distance is less effective than the
digit-based objectives in the targeted setting, achieving
$15.73\%$ and $27.59\%$, whereas Digit~CE achieves
$35.24\%$ and $50.69\%$, respectively.
As discussed in Sec.~\ref{sec:missclick_t}, moving the
expected digit values toward the target does not guarantee
that the target digits become the argmax.
This indicates that directly optimizing target-digit
probabilities better matches the targeted attack goal than
Soft-coordinate distance.
Under a common optimization procedure, the two attack goals
favor different objectives: click displacement for
untargeted disruption and discrete target-digit generation
for targeted redirection.
This result supports the goal-specific execution-aware
design of MissClick-U and MissClick-T.
Appendix~\ref{sec:supp_analysis} further relates the
teacher-forced optimization signals to the final
greedy-generated clicks.
Appendix~\ref{sec:supp_digit_position} reports how
place-value weighting changes target-digit probabilities
and generated-digit match rates at the hundreds, tens, and
ones positions.

\subsection{Effect of Place-Value Weighting}
\label{sec:place_ablation}

We compare Digit~CE, which assigns equal weight to all
digit positions, with Place-weighted digit~CE, which assigns
weights according to decimal place values.
All other attack settings remain identical, and the
place-weighted loss is normalized by the total place weight.
According to the targeted results reported in
Table~\ref{tab:objective_analysis}, Place-weighted digit~CE
outperforms Digit~CE on both models and all three platforms.
The overall targeted ASR improves from $35.24\%$ to
$44.86\%$ on OS-Atlas, corresponding to an improvement of
$9.62$ percentage points, and from $50.69\%$ to $62.67\%$
on UGround, corresponding to an improvement of $11.98$
percentage points.
These results show that, within the targeted digit
objective, accounting for the non-uniform geometric effects
of digit positions through place-value weighting improves
attack effectiveness compared with uniform weighting.

\subsection{Further Analysis}
\label{sec:analysis}

\begin{wraptable}{r}{0.49\columnwidth}
\vspace{-0.8\baselineskip}
\centering
\small
\vspace{-0.3em}
\caption{
Attack success rates under different perturbation budgets
$\varepsilon$ on separate fixed stratified subsets of
300 tasks for the untargeted and targeted settings.
}
\vspace{-0.6em}
\setlength{\dashlinedash}{2pt}
\setlength{\dashlinegap}{2pt}
\setlength{\tabcolsep}{3pt}

\resizebox{\linewidth}{!}{
\begin{tabular}{|c|ccc:ccc|}
\hline

\multirow{3}{*}{Method}
& \multicolumn{6}{c|}{Untargeted Attacks} \\
\cline{2-7}

& \multicolumn{3}{c:}{OS-Atlas}
& \multicolumn{3}{c|}{UGround} \\
\cline{2-7}

& $\varepsilon{=}4$ & $\varepsilon{=}8$ & $\varepsilon{=}16$
& $\varepsilon{=}4$ & $\varepsilon{=}8$ & $\varepsilon{=}16$ \\
\hline

Representation
& 33.73\% & 47.39\% & 55.42\%
& 12.45\% & 29.06\% & 39.25\% \\

\hdashline
\rowcolor{gray!15}
\textbf{MissClick-U}
& \textbf{54.62\%} & \textbf{69.88\%} & \textbf{75.90\%}
& \textbf{54.34\%} & \textbf{64.53\%} & \textbf{74.72\%} \\

\hline

\multirow{3}{*}{Method}
& \multicolumn{6}{c|}{Targeted Attacks} \\
\cline{2-7}

& \multicolumn{3}{c:}{OS-Atlas}
& \multicolumn{3}{c|}{UGround} \\
\cline{2-7}

& $\varepsilon{=}4$ & $\varepsilon{=}8$ & $\varepsilon{=}16$
& $\varepsilon{=}4$ & $\varepsilon{=}8$ & $\varepsilon{=}16$ \\
\hline

Token CE
& 3.72\% & 11.52\% & 13.75\%
& 5.47\% & 11.68\% & 17.15\% \\

\hdashline
\rowcolor{red!10}
\textbf{MissClick-T}
& \textbf{24.91\%} & \textbf{40.15\%} & \textbf{46.47\%}
& \textbf{34.67\%} & \textbf{55.84\%} & \textbf{66.06\%} \\
\hline
\end{tabular}
}
\label{tab:eps}
\end{wraptable}

\textbf{Perturbation-budget sensitivity.}
We evaluate MissClick and the corresponding baselines under
$\varepsilon\in\{4,8,16\}/255$ on fixed stratified subsets
of 300 untargeted and targeted tasks, with all other settings
unchanged.
Table~\ref{tab:eps} reports the micro-averaged ASR over the
three platforms.
MissClick outperforms the baselines at every tested budget.
Notably, MissClick-U at $\varepsilon=4/255$ exceeds
Representation attack at $\varepsilon=16/255$ on UGround
($54.34\%$ vs.\ $39.25\%$), while MissClick-T shows the
same cross-budget advantage on both models
($24.91\%$ vs.\ $13.75\%$ and $34.67\%$ vs.\ $17.15\%$).

\begin{wrapfigure}{r}{0.49\columnwidth}
  \vspace{-0.8\baselineskip}
  \centering

  \begin{subfigure}[t]{0.49\linewidth}
    \centering
    \includegraphics[width=\linewidth]
    {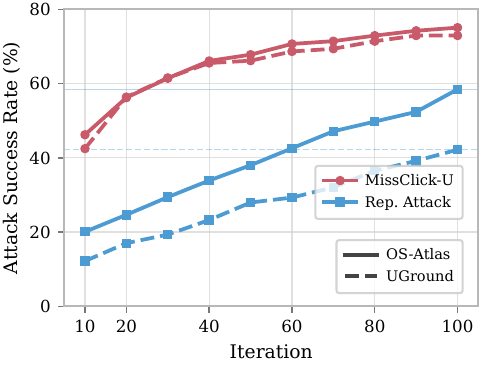}
    \caption{Untargeted attacks.}
    \label{fig:iteration_untargeted}
  \end{subfigure}
  \hfill
  \begin{subfigure}[t]{0.49\linewidth}
    \centering
    \includegraphics[width=\linewidth]
    {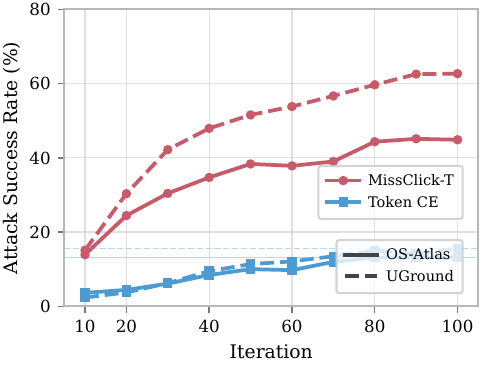}
    \caption{Targeted attacks.}
    \label{fig:iteration_targeted}
  \end{subfigure}

  \vspace{-0.6em}
  \caption{
  Attack success rate under different optimization iteration
  budgets.
  The horizontal lines indicate the 100-iteration ASR of the
  corresponding baselines.
  Solid and dashed curves denote OS-Atlas and UGround,
  respectively.
  }
  \label{fig:asr_iteration}
  \vspace{-0.8\baselineskip}
\end{wrapfigure}
\textbf{Iteration-budget analysis.} 
Under the same attack settings, we plot the micro-averaged
ASR against the number of optimization iterations.
The results are shown in Fig.~\ref{fig:asr_iteration}, where
the horizontal lines indicate the corresponding baseline
ASR after 100 iterations.
MissClick reaches or exceeds these reference results within
10--30 iterations in all four settings.
On UGround, MissClick-U achieves $42.48\%$ ASR after
10 iterations, compared with $42.21\%$ for Representation
attack after 100 iterations.
This indicates that MissClick requires fewer optimization
iterations to reach the corresponding baseline performance
and achieves stronger attack performance under limited
iteration budgets.

\noindent\textbf{Qualitative examples.}
Figure~\ref{fig:overview} illustrates an executable targeted
click redirection.
Additional untargeted and targeted examples across both
models and all three platforms are provided in
Appendix~\ref{sec:supp_qualitative}, together with
perturbation visualizations under different budgets.

\subsection{Cross-Model Transfer and Input Preprocessing}
\label{sec:transfer_preprocessing}

\textbf{Cross-model transferability.}
We evaluate transfer between OS-Atlas and UGround without
re-optimization.
As reported in Appendix Table~\ref{tab:supp_transfer},
MissClick-U obtains untargeted transfer ASRs of $17.01\%$
from OS-Atlas to UGround and $38.56\%$ in the reverse
direction, compared with $14.16\%$ and $54.72\%$ for
Representation attack.
MissClick-T achieves targeted transfer ASRs of $3.37\%$ and
$8.71\%$, compared with $1.75\%$ and $4.16\%$ for Token~CE.
Overall, cross-model transfer remains substantially weaker
than white-box performance and varies with transfer
direction and attack objective.

\textbf{Non-adaptive input preprocessing.}
We evaluate fixed JPEG compression and Gaussian blur without
incorporating either transformation into attack optimization.
Across both models, these transformations reduce MissClick-U
ASR to $13.09\%$--$20.92\%$ and MissClick-T ASR to
$2.25\%$--$4.55\%$.
They also reduce clean grounding accuracy by up to $4.95$
percentage points.
These results show that fixed preprocessing substantially
suppresses MissClick but also affects clean grounding accuracy. 
Complete results are reported in
Appendix Table~\ref{tab:supp_defense}.
\section{Related Work}
\label{sec:related_work}

\textbf{GUI visual grounding.}
GUI visual grounding maps natural-language instructions to
screen regions~\citep{cheng2024seeclick}.
Many methods generate point or bounding-box coordinates
autoregressively~\citep{cheng2024seeclick,gou2025uground,wu2025osatlas}.
UGround~\citep{gou2025uground} targets cross-platform
generalization, while OS-Atlas~\citep{wu2025osatlas} uses a
unified multi-platform action space.
GUI-Actor~\citep{wu2025gui} instead uses a coordinate-free
action head.
MissClick targets coordinate-generating models, where digit
tokens are parsed into executable clicks.

\textbf{Adversarial perturbation attacks.}
Adversarial examples are well studied in image
classification~\citep{goodfellow2015explaining,
carlini2017towards,croce2020reliable}.
For GUI grounding, \citet{zhao2025robustness}
optimize visual-embedding distance or target-coordinate
probability.
Related autoregressive attacks optimize token-sequence
objectives~\citep{carlini2023aligned,bailey2024image}.
These objectives operate before numerical parsing and click
mapping, whereas MissClick directly targets click-level
outcomes.
Our setting is complementary to prompt injection, backdoors,
and broader agent attacks~\citep{wang2025webinject,ye2025visualtrap,
wu2025dissecting,zhang2025realistic,
debenedetti2024agentdojo}.

\textbf{Numerical-aware objectives.}
Prior supervised objectives incorporate numerical structure
during training.
Number Token Loss models numerical distance
~\citep{zausinger2025regress}, while 
~\citet{fei2025advancing} combine Earth Mover's Distance with
position weighting.
DIST\textsuperscript{2}Loss uses token-distance-based soft
targets~\citep{chung2025teaching}, and Phi-Ground studies
coordinate-digit smoothing and position weighting
~\citep{zhang2025phiground}.
Unlike these training objectives, MissClick keeps the victim
model fixed and optimizes a bounded screenshot perturbation:
MissClick-U maximizes soft-click displacement, whereas
MissClick-T optimizes place-weighted target digits.
\section{Conclusion, Limitations, and Future Work}
\label{sec:conclusion}

This work identifies an execution-level attack surface at
the interface between coordinate generation and click
execution in GUI grounding models.
MissClick aligns attack optimization with this interface
through soft-coordinate displacement for untargeted
disruption and place-weighted target-digit optimization for
targeted redirection.
Across two models and three platforms, MissClick-U improves
untargeted ASR by $16.62$ and $30.72$ percentage points,
while MissClick-T improves targeted ASR by $31.73$ and
$47.06$ percentage points over the corresponding baselines.
The objective comparison further shows that untargeted
and targeted attacks favor different objectives.
Our evaluation covers two models under a white-box threat model, both of which generate per-digit decimal coordinates on ScreenSpot-v2. MissClick-T also treats all incorrect values within each digit position uniformly.
Cross-model transfer remains weaker than white-box
performance, while fixed input preprocessing reduces both
attack success and clean grounding accuracy.
Future work will extend execution-aware attacks to broader
output interfaces, ordinal-aware objectives, query-based
settings, and adaptive defenses.

\subsection*{AI use statement}
Generative AI tools were used to assist with manuscript
preparation, primarily for improving English expression,
clarity, organization, and readability.
The authors reviewed and edited all AI-assisted content and
independently verified the technical claims, equations,
citations, experimental results, and conclusions presented
in the paper.
Generative AI tools were not used to generate experimental
data or replace the authors' technical judgment.
The authors take full responsibility for the final content
of the manuscript.

\subsection*{Ethics statement}

This work studies adversarial perturbations against GUI
grounding models and therefore has potential dual-use
implications.
The proposed techniques are intended to identify
vulnerabilities in coordinate-generating GUI grounding
systems and to support the development of more robust models
and execution safeguards.
All experiments were conducted in an offline research
setting using publicly available models and benchmark data.
The study does not involve human participants, private user
data, or interactions with deployed third-party systems.
We acknowledge that adversarial techniques may potentially
be misused, and we present the results for the purpose of
security analysis and robustness research.

\subsection*{Reproducibility statement}

We provide detailed descriptions of the threat model,
generation-to-execution pipeline, attack objectives, and
optimization procedure in Sec.~\ref{sec:method}.
The experimental setup, including victim models, dataset,
baselines, attack settings, evaluation metrics, and decoding
procedure, is described in Sec.~\ref{sec:experiments}.
Additional implementation details, model-specific coordinate
formats, coordinate parsing procedures, evaluation protocols,
and the complete attack algorithm are provided in the
appendix.
The appendix also reports additional experimental results
and analyses to facilitate verification and reproduction of
the reported findings.

\bibliography{yyyy2027_conference}
\bibliographystyle{yyyy2027_conference}

\appendix

\section{Complete Attack Procedure and Implementation Details}
\label{sec:supp_impl_eval}

This section provides the complete MissClick optimization
procedure and the model-specific details used to instantiate
its execution-aware objectives.
Algorithm~\ref{alg:missclick} shows how MissClick-U and
MissClick-T share the same projected-gradient optimization
procedure while using different reference coordinates and
attack objectives.
The remainder of this section describes the native input and
output formats of the victim models, coordinate-digit
alignment, objective implementations, and attack settings.
\begin{algorithm}[h]
\caption{MissClick optimization with goal-specific
execution-aware objectives}
\label{alg:missclick}
\begin{algorithmic}[1]

\REQUIRE GUI grounding model $\mathcal{M}$, screenshot
$\mathbf{I}$, instruction $q$, reference coordinates
$c^{\mathrm{ref}}$, perturbation budget $\varepsilon$,
step size $\alpha$, number of iterations $N$, and attack
mode $m\in\{\mathrm{U},\mathrm{T}\}$

\ENSURE Adversarial screenshot
$\widetilde{\mathbf{I}}$

\STATE Set $c^{\mathrm{ref}}\leftarrow c^{\mathrm{gt}}$
if $m=\mathrm{U}$; otherwise set
$c^{\mathrm{ref}}\leftarrow c^{\mathrm{tgt}}$

\STATE Construct a model-native reference answer sequence
containing $c^{\mathrm{ref}}$ for teacher forcing

\STATE Extract coordinate-digit positions $(k,j)$ and
place values $a_{k,j}$ from the reference answer sequence

\STATE Initialize $\boldsymbol{\delta}$ randomly within
$[-\varepsilon,\varepsilon]^d$

\FOR{$i=1,\ldots,N$}

    \STATE Run a teacher-forced forward pass on
    $(\mathbf{I}+\boldsymbol{\delta},q)$ using the
    reference answer sequence and the frozen model
    $\mathcal{M}$

    \STATE Extract digit distributions $p_{k,j}(v)$
    using \eqref{eq:digit_distribution}

    \IF{$m=\mathrm{U}$}

        \STATE Compute soft coordinates $\widetilde{c}$
        using \eqref{eq:expected_digit}
        and~\eqref{eq:soft_coord}

        \STATE
        $\mathcal{L}\leftarrow\mathcal{L}_{\mathrm{U}}$
        using \eqref{eq:untargeted_loss}

        \STATE
        $\boldsymbol{\delta}\leftarrow
        \Pi_{[-\varepsilon,\varepsilon]^d}
        \left(
        \boldsymbol{\delta}
        +\alpha\,
        \operatorname{sign}
        \left(
        \nabla_{\boldsymbol{\delta}}\mathcal{L}
        \right)
        \right)$
        \COMMENT{maximize click displacement}

    \ELSIF{$m=\mathrm{T}$}

        \STATE
        $\mathcal{L}\leftarrow\mathcal{L}_{\mathrm{T}}$
        using \eqref{eq:targeted_loss}

        \STATE
        $\boldsymbol{\delta}\leftarrow
        \Pi_{[-\varepsilon,\varepsilon]^d}
        \left(
        \boldsymbol{\delta}
        -\alpha\,
        \operatorname{sign}
        \left(
        \nabla_{\boldsymbol{\delta}}\mathcal{L}
        \right)
        \right)$
        \COMMENT{minimize target-digit loss}

    \ENDIF

\ENDFOR

\RETURN
$\widetilde{\mathbf{I}}
=
\operatorname{clamp}
\left(
\mathbf{I}+\boldsymbol{\delta},0,1
\right)$

\end{algorithmic}
\end{algorithm}

\subsection{Complete Attack Procedure}
\label{sec:supp_algorithm}

MissClick keeps the victim model fixed and updates only the
screenshot perturbation.
At each iteration, a teacher-forced forward pass provides
the coordinate-digit distributions used by the selected
execution-aware objective.
MissClick-U maximizes soft-click displacement from the
ground-truth click, whereas MissClick-T minimizes the
place-weighted loss of the target digits.

\renewcommand{\algorithmicrequire}{\textbf{Input:}}
\renewcommand{\algorithmicensure}{\textbf{Output:}}

Teacher forcing is used only to obtain differentiable
optimization signals.
Attack success is determined from the coordinates produced
by standard greedy autoregressive generation without the
reference answer sequence.

\subsection{Model-Specific Inputs and Coordinate Outputs}

\label{sec:supp_model_io}

\paragraph{Prompts and Native Coordinate Formats.}
\label{par:supp_prompts_formats}
We evaluate OS-Atlas-Base-7B and UGround-V1-7B using their
respective native grounding prompts and coordinate formats.
For OS-Atlas, the user prompt is
\begin{quote}
\small\ttfamily
In this UI screenshot, what is the position of the element
corresponding to the command ``$q$'' (with bbox)?
\end{quote}
where $q$ denotes the task instruction. OS-Atlas generates a
bounding-box coordinate
$c=(x_1,y_1,x_2,y_2)$ in the normalized $[0,1000]$
coordinate system. Its native supervised answer contains an
object-reference segment followed by a bounding-box segment of
the form
\begin{quote}
\small
\texttt{<|object\_ref\_start|>}$q$\texttt{<|object\_ref\_end|>}
\texttt{<|box\_start|>}$(x_1,y_1),(x_2,y_2)$
\texttt{<|box\_end|>}.
\end{quote}
For UGround, the user prompt is
\begin{quote}
\small\ttfamily
Your task is to help the user identify the precise coordinates
(x, y) of a specific area/element/object on the screen based on
a description.\\
Description: $q$\\
Answer:
\end{quote}
UGround directly generates a point coordinate
$c=(x,y)$ in the same normalized coordinate system. Its native
reference answer is represented as $(x,y)$.
The two native formats are converted to executed clicks using
the mapping $r(\cdot)$ defined in the main paper. For an
OS-Atlas bounding box, we use its center,
\[
r(c)=
\left(
\frac{x_1+x_2}{2},
\frac{y_1+y_2}{2}
\right).
\]
For a UGround point output, we directly use
$r(c)=(x,y)$.
\paragraph{Reference Answers and Coordinate-Digit Alignment.}
\label{par:supp_reference_alignment}
During attack optimization, each input is paired with a
model-native reference answer. For untargeted attacks, the
reference answer contains the ground-truth coordinate output
$c^{\mathrm{gt}}$. For targeted attacks, the task instruction
and screenshot remain unchanged, while the coordinate part of
the reference answer is replaced with the attacker-specified
target coordinate output $c^{\mathrm{tgt}}$.
For OS-Atlas, $c^{\mathrm{gt}}$ or $c^{\mathrm{tgt}}$ is
represented as a bounding box enclosed by the native
\texttt{<|box\_start|>} and \texttt{<|box\_end|>} markers.
For UGround, the corresponding reference is represented as a
point coordinate. The targeted UGround reference uses the
model-native point associated with the center of the selected
target region. The final targeted success criterion nevertheless
uses the complete target region $B_{\mathrm{tgt}}$ rather than
requiring exact equality with this point coordinate.
The supervised answer positions are identified from the
teacher-forced labels, where prompt and padding positions are
masked with $-100$. We then locate the individual digit tokens
within the coordinate span of the reference answer. For each
coordinate component $c_k$, the $j$-th digit is associated with
its token position $t_{k,j}$ and decimal place value
$a_{k,j}=10^{m_k-j}$, where $m_k$ is the number of digits in
$c_k$. We therefore record each coordinate digit as
$(t_{k,j}, k, a_{k,j})$.
Here, $t_{k,j}$ denotes its absolute position in the complete
teacher-forced token sequence, $k$ identifies the coordinate
component, and $a_{k,j}$ specifies its numerical place value.
Since the logits at position $t_{k,j}-1$ predict the token at
$t_{k,j}$, these indices are used to extract the corresponding
next-token digit logits.
For OS-Atlas, we identify the four coordinate components within
the native bounding-box markers and assign them to $x_1$, $y_1$,
$x_2$, and $y_2$ in order. For UGround, we identify the two
coordinate components in its native point-format answer and assign
them to $x$ and $y$. 
We then identify the token position of each coordinate digit
in the teacher-forced reference answer. Each position is
associated with its coordinate component and decimal place
value.

\begin{table*}[t]
    \centering
    \small
    \caption{Core attack settings.}
    \label{tab:supp_runtime_configuration}
    \begin{tabular}{ll}
        \hline
        Item & Configuration \\
        \hline
        Victim models & OS-Atlas-Base-7B; UGround-V1-7B \\
        Max pixels & 1024 \\
        Perturbation budget & $16/255$ \\
        Step size & $1/255$ \\
        Iterations (optimized methods) & 100 \\
        \hline
    \end{tabular}
\end{table*}

\subsection{Attack Optimization Details}
\label{sec:supp_attack_implementation}
\paragraph{Coordinate-Digit Identification and Logit Alignment.}
\label{par:supp_digit_logit_alignment}
A causal language model uses the logits at position
$t_{k,j}-1$ to predict the token at position $t_{k,j}$.
Therefore, for each coordinate-digit position identified above,
the implementation extracts the corresponding next-token logit
vector. It then selects the logits associated with the decimal
digits \texttt{0}, \ldots, \texttt{9} using the victim model's
tokenizer. These digit-level logits are used to construct the
probability distributions and attack objectives described below.
\paragraph{Digit Distributions and Soft-Coordinate Reconstruction.}
\label{par:supp_soft_coordinate_impl}
Let $z_{k,j,v}$ denote the next-token logit for digit $v$ at
position $t_{k,j}-1$. A softmax over the ten digit logits produces the
digit-restricted distribution
$p_{k,j}(v)$:
\begin{equation}
p_{k,j}(v)
=
\frac{\exp(z_{k,j,v})}
{\sum_{u=0}^{9}\exp(z_{k,j,u})},
\qquad v\in\{0,\ldots,9\}.
\end{equation}
The expected digit value is computed as
\[
\widetilde d_{k,j}
=
\sum_{v=0}^{9} v\,p_{k,j}(v),
\]
and each soft coordinate component is reconstructed by
\[
\widetilde c_k
=
\sum_{j=1}^{m_k}
a_{k,j}\widetilde d_{k,j}.
\]
For OS-Atlas, the four reconstructed components form
$\widetilde c=(\widetilde x_1,\widetilde y_1,
\widetilde x_2,\widetilde y_2)$, and the soft click is the
center of this soft box. For UGround, the two reconstructed
components directly form the soft click
$(\widetilde x,\widetilde y)$.
The reconstructed soft click is used by the soft-coordinate
objective and is later compared with the final generated click.
This computation uses only the digit-restricted distributions
during optimization. During final evaluation, the model generates its response from
the full vocabulary without digit-only or format-constrained
decoding.

\paragraph{Baselines and Objective Variants.}
\label{par:supp_baseline_variants}
Except for Random noise, the compared methods use the same
perturbation budget and optimization settings but differ in
their loss functions.

\emph{Random noise} uses a random  perturbation clipped to the
same pixel-space $\ell_\infty$ budget, without gradient updates.

\emph{Token CE} applies standard cross-entropy to all supervised
tokens in the model-native reference answer. The loss is maximized
on the ground-truth answer for untargeted attacks and minimized
on the target answer for targeted attacks.

\emph{Representation attack} maximizes the mean-squared
difference between the clean and adversarial visual
embeddings. Because this objective does not specify a target
coordinate or region, it is used only for untargeted attacks.

\emph{Digit CE} applies cross-entropy only at the coordinate-digit
positions and assigns equal weight to every digit. The digit loss
is maximized for untargeted attacks and minimized for targeted
attacks.

\emph{Soft-coordinate distance} optimizes the squared distance
between the reconstructed soft click and a reference click.
For untargeted attacks, MissClick-U maximizes this distance from
the ground-truth click. For targeted attacks, Soft-coordinate
distance minimizes the distance to the center of the target
region.

\emph{Place-weighted digit CE} weights the cross-entropy at each
coordinate-digit position by its decimal place value and
normalizes the result by the total place weight. For untargeted
attacks, it maximizes the weighted loss on the ground-truth
digits. For targeted attacks, MissClick-T minimizes the weighted
loss on the target digits.
\begin{table*}[t]
    \centering
    \small
\caption{Numbers of evaluated samples after model-specific
clean-correct filtering and, for targeted attacks, valid-target
filtering.}
    \label{tab:supp_sample_counts}
    \begin{tabular}{llrrrr}
        \hline
        Setting & Model & Desktop & Web & Mobile & All \\
        \hline
        Untargeted & OS-Atlas & 273 & 362 & 436 & 1,071 \\
        Untargeted & UGround  & 295 & 372 & 456 & 1,123 \\
        Targeted   & OS-Atlas & 152 & 271 & 346 & 769 \\
        Targeted   & UGround  & 161 & 279 & 361 & 801 \\
        \hline
    \end{tabular}
\end{table*}

\begin{table*}[t]
    \centering
    \small
    \setlength{\tabcolsep}{3.5pt}
    \caption{Parse validity of adversarial outputs on
    ScreenSpot-v2. Untargeted outputs are evaluated on
    model-specific clean-correct sets (1071 for OS-Atlas,
    1123 for UGround). Targeted outputs are evaluated on
    model-specific clean-correct targeted sets (769 for
    OS-Atlas, 801 for UGround). Invalid outputs count as
    untargeted successes and targeted failures.}
    \label{tab:supp_parse_validity}
    \begin{tabular}{lllrrrc}
        \toprule
        Setting & Model & Method
        & Outputs & Valid & Invalid & Valid rate \\
        \midrule
        Untargeted
        & OS-Atlas & Random noise & 1071 & 1071 & 0 & 100.00\% \\
        &  & Token CE & 1071 & 1068 & 3 & 99.72\% \\
        &  & Representation attack & 1071 & 1071 & 0 & 100.00\% \\
        &  & Digit CE & 1071 & 1070 & 1 & 99.91\% \\
        &  & Place-weighted digit CE & 1071 & 1070 & 1 & 99.91\% \\
        &  & MissClick-U & 1071 & 1071 & 0 & 100.00\% \\
        \cmidrule{2-7}
        & UGround & Random noise & 1123 & 1123 & 0 & 100.00\% \\
        &  & Token CE & 1123 & 1123 & 0 & 100.00\% \\
        &  & Representation attack & 1123 & 1123 & 0 & 100.00\% \\
        &  & Digit CE & 1123 & 1123 & 0 & 100.00\% \\
        &  & Place-weighted digit CE & 1123 & 1123 & 0 & 100.00\% \\
        &  & MissClick-U & 1123 & 1123 & 0 & 100.00\% \\
        \midrule
        Targeted
        & OS-Atlas & Token CE & 769 & 768 & 1 & 99.87\% \\
        &  & Soft-coordinate distance & 769 & 769 & 0 & 100.00\% \\
        &  & Digit CE & 769 & 769 & 0 & 100.00\% \\
        &  & MissClick-T & 769 & 769 & 0 & 100.00\% \\
        \cmidrule{2-7}
        & UGround & Token CE & 801 & 801 & 0 & 100.00\% \\
        &  & Soft-coordinate distance & 801 & 801 & 0 & 100.00\% \\
        &  & Digit CE & 801 & 801 & 0 & 100.00\% \\
        &  & MissClick-T & 801 & 801 & 0 & 100.00\% \\
        \bottomrule
    \end{tabular}
\end{table*}
\begin{table*}[t]
    \centering
    \small
    \setlength{\tabcolsep}{3.5pt}
\caption{Target area, source-to-target center distance, and
source--target IoU in the common 896-task target-construction
set. Areas and distances are computed from normalized $[0,1]$
coordinates. Here, $n$ is the number of tasks; Min and Max are
the minimum and maximum; Q1, Median, and Q3 are the 25th, 50th,
and 75th percentiles; and IQR equals Q3 minus Q1. Target-area values are reported as percentages of the screen area.}
    \label{tab:supp_target_statistics}
    \begin{tabular}{llrrrrrrr}
        \toprule
        Platform & Quantity & $n$ & Min & Q1 & Median
        & Q3 & IQR & Max \\
        \midrule
Desktop & Target area
& 178 & 0.068\% & 0.354\% & 0.729\%
& 1.579\% & 1.225\% & 52.417\% \\
        & Center distance
        & 178 & 0.022663 & 0.199514 & 0.391831
        & 0.630893 & 0.431379 & 1.202555 \\
        & Source--target IoU
        & 178 & 0 & 0 & 0 & 0 & 0 & 0 \\
        \addlinespace

Web & Target area
& 323 & 0.018\% & 0.170\% & 0.385\%
& 1.024\% & 0.854\% & 22.973\% \\
        & Center distance
        & 323 & 0.031281 & 0.248606 & 0.404980
        & 0.602149 & 0.353544 & 1.211590 \\
        & Source--target IoU
        & 323 & 0 & 0 & 0 & 0 & 0 & 0 \\
        \addlinespace

Mobile & Target area
& 395 & 0.084\% & 0.598\% & 1.399\%
& 3.932\% & 3.334\% & 15.990\% \\
        & Center distance
        & 395 & 0.069027 & 0.282592 & 0.438890
        & 0.665074 & 0.382481 & 1.066478 \\
        & Source--target IoU
        & 395 & 0 & 0 & 0 & 0 & 0 & 0 \\
        \addlinespace

All & Target area
& 896 & 0.018\% & 0.353\% & 0.798\%
& 2.173\% & 1.820\% & 52.417\% \\
        & Center distance
        & 896 & 0.022663 & 0.254889 & 0.422449
        & 0.634902 & 0.380014 & 1.211590 \\
        & Source--target IoU
        & 896 & 0 & 0 & 0 & 0 & 0 & 0 \\
        \bottomrule
    \end{tabular}
\end{table*}
\begin{table*}[t]
    \centering
    \small
    \setlength{\tabcolsep}{3.5pt}
\caption{Targeted attack success by ScreenSpot-v2 element type.
Each cell reports successes/tasks and ASR.}
    \label{tab:supp_element_type}
\begin{tabular}{lcccc}
    \toprule
    Objective & OS-Atlas text & OS-Atlas icon/widget
    & UGround text & UGround icon/widget \\
    \midrule
    MissClick-T
    & 183/469; 39.0\%
    & 162/300; 54.0\%
    & 294/481; 61.1\%
    & 208/320; 65.0\% \\
    Digit CE
    & 142/469; 30.3\%
    & 129/300; 43.0\%
    & 235/481; 48.9\%
    & 171/320; 53.4\% \\
    Soft-coordinate distance
    & 59/469; 12.6\%
    & 62/300; 20.7\%
    & 116/481; 24.1\%
    & 105/320; 32.8\% \\
    Token~CE
    & 31/469; 6.6\%
    & 70/300; 23.3\%
    & 53/481; 11.0\%
    & 72/320; 22.5\% \\
    \bottomrule
\end{tabular}
\end{table*}

\paragraph{Attack Settings.}
\label{par:supp_matched_settings}

The core attack settings are summarized in
Table~\ref{tab:supp_runtime_configuration}. Within each model and
attack setting, the compared objectives use the same screenshots,
instructions, model-specific prompts, reference-answer
construction, preprocessing, perturbation constraint, random
seed, and evaluator. During optimization, the victim-model
parameters are frozen and only the image perturbation is updated.
Except for Random noise, all methods update the image perturbation
using gradients and constrain it to the same $\ell_\infty$ budget
after every step. The final saved images are evaluated using the
same autoregressive generation, coordinate parser, click mapping,
and success criteria. Reported ASR values are computed on the same
model-specific clean-correct task sets.

\subsection{Teacher-Forced Optimization and Greedy Autoregressive Evaluation}
\label{sec:supp_teacher_greedy}
\paragraph{Teacher-Forced Optimization Path.}
\label{par:supp_teacher_path}
Teacher forcing is used only to construct a differentiable
optimization path. The input sequence contains the screenshot,
task instruction, and model-native reference answer. The
reference answer contains $c^{\mathrm{gt}}$ for an untargeted
attack and $c^{\mathrm{tgt}}$ for a targeted attack. The victim
model remains frozen.
The forward pass yields differentiable next-token logits at the
fixed reference-answer positions. MissClick extracts the logits
at coordinate-digit positions, forms digit-restricted
probability distributions, and computes either the soft-coordinate
objective or the target-digit objective. Gradients are propagated
from this objective to the screenshot perturbation.
The reference answer, teacher-forced logits, soft coordinates,
and monitored digit predictions are used only during attack
optimization. They are not used to determine whether the final
attack succeeds.
\paragraph{Final Generation and Evaluation Path.}
\label{par:supp_greedy_path}
After optimization, the final adversarial screenshot is saved
and evaluated in a separate inference pass. The reference answer
is not provided in this pass. The model receives only the
adversarial screenshot and the original task instruction and
performs greedy autoregressive generation over the complete
vocabulary. The generated textual response is parsed using the
model-specific coordinate parser, mapped to an executed click,
and compared with $B_{\mathrm{gt}}$ or $B_{\mathrm{tgt}}$.
All ASR values in the main paper are computed from this path.
\section{Evaluation Protocol and Target Construction}
\label{sec:supp_eval_protocol}
This section specifies how evaluation samples are constructed
and how freely generated textual outputs are converted into
attack success labels. The same protocol is used for all methods
under each attack setting.
\subsection{Clean-Correct Filtering and ASR Computation}
\label{sec:supp_filtering_asr}
\paragraph{Evaluation Sample Construction.}
\label{par:supp_sample_construction}
ScreenSpot-v2 contains $1{,}272$ tasks: $334$ desktop,
$437$ web, and $501$ mobile tasks. Each task contains a
screenshot, a natural-language instruction, and a ground-truth
element bounding box.
We first run each victim model on the unperturbed screenshots
using its native prompt and output format. A task is considered
clean-correct if the freely generated response can be parsed and
the resulting executed click falls inside the ground-truth region
$B_{\mathrm{gt}}$. Clean correctness is determined independently
for OS-Atlas and UGround because the models can succeed on
different subsets of tasks.
The untargeted evaluation set for a victim model contains all of
its clean-correct tasks. This gives $1{,}071$ samples for
OS-Atlas and $1{,}123$ samples for UGround.
For targeted attacks, we first construct a common set of $896$
ScreenSpot-v2 tasks that contain at least one valid target
element. For each victim model, the targeted evaluation set includes only
tasks that have a valid target element and are clean-correct for
that model. This gives $769$ samples for
OS-Atlas and $801$ samples for UGround.
The same model-specific clean-correct set is used for all attack
methods compared under the corresponding setting.
\paragraph{Denominators and Micro-Averaging.}
\label{par:supp_denominators_micro}

For a model $\mathcal M$ and attack setting $m$, ASR is defined
as
\[
\operatorname{ASR}
=
\frac{
\sum_{i\in\mathcal E_{\mathcal M,m}}
\mathbb I[\operatorname{Success}(i)]
}{
|\mathcal E_{\mathcal M,m}|
},
\]
where $\mathcal E_{\mathcal M,m}$ is the corresponding
model- and setting-specific evaluation set defined above.
For untargeted attacks,
$\operatorname{Success}(i)=1$ when the executed click falls
outside $B_{\mathrm{gt}}$. For targeted attacks,
$\operatorname{Success}(i)=1$ when the executed click falls
inside $B_{\mathrm{tgt}}$.
The ``All'' result is a micro-average. We first sum successful
attacks over desktop, web, and mobile and then divide by the
total number of evaluated samples:
\[
\operatorname{ASR}_{\mathrm{All}}
=
\frac{
n_{\mathrm{succ}}^{\mathrm{desktop}}
+
n_{\mathrm{succ}}^{\mathrm{web}}
+
n_{\mathrm{succ}}^{\mathrm{mobile}}
}{
n^{\mathrm{desktop}}
+
n^{\mathrm{web}}
+
n^{\mathrm{mobile}}
}.
\]
It is not obtained by simply averaging the three
platform-specific ASR values. Table~\ref{tab:supp_sample_counts}
reports the corresponding sample counts by setting, model, and
platform.


\subsection{Coordinate Parsing and Executed-Click Evaluation}
\label{sec:supp_parsing_click}
\paragraph{Parsing and Click Mapping.}
\label{par:supp_parsing_mapping}
For OS-Atlas, the evaluator extracts four numerical values from the
native bounding-box response and forms
\[
c=(x_1,y_1,x_2,y_2).
\]
The executed click is the center of the predicted box. For
UGround, it extracts the point coordinate
$c=(x,y)$ and directly uses this point as the executed click.
Before checking whether a click falls inside a bounding box,
the generated $[0,1000]$ click coordinates are normalized to
$[0,1]$ to match the bounding boxes in the evaluation JSON files.
A point on the boundary of an annotated region is considered
inside the region. Thus, a click
$\mathbf u=(u_x,u_y)$ lies in
$B=(x_1,y_1,x_2,y_2)$ when
\[
x_1\leq u_x\leq x_2
\quad\text{and}\quad
y_1\leq u_y\leq y_2.
\]
\paragraph{Invalid-Output Handling.}
\label{par:supp_invalid_outputs}
An output is invalid when the formal evaluator cannot convert
the generated response into the required numerical coordinate
format.
If the model output on the clean screenshot cannot be parsed,
the task is excluded because it cannot be verified as
clean-correct.
In the untargeted setting, an unparseable response does not
produce a valid click inside $B_{\mathrm{gt}}$ and is counted
as an untargeted attack success. In the targeted setting, an
unparseable response does not produce an executed click inside
$B_{\mathrm{tgt}}$ and is counted as a targeted attack failure.

Table~\ref{tab:supp_parse_validity} reports parse validity
for every evaluated model and attack objective.
In the untargeted setting, $13{,}159$ of $13{,}164$
outputs are parseable, corresponding to a validity rate of
$99.96\%$.
The five invalid outputs all occur on OS-Atlas:
Token~CE produces three, Digit~CE produces one, and
Place-weighted digit~CE produces one.
Every untargeted output generated by UGround is parseable,
and MissClick-U achieves a $100\%$ valid rate on both
victim models.

In the targeted setting, $6{,}279$ of $6{,}280$ outputs
are parseable, corresponding to a validity rate of
$99.98\%$.
The only invalid targeted output is produced by OS-Atlas
Token~CE.
MissClick-T achieves a $100\%$ valid rate on both victim
models.
Consequently, every success reported for MissClick-U and
MissClick-T corresponds to an executable parsed click
rather than an output-format failure.

\begin{table*}[t]
    \centering
    \small
    \setlength{\tabcolsep}{3.5pt}
    \caption{Targeted ASR by fixed target-area quartile.
    Each cell reports successes/tasks and ASR.
    The two model panels use the same frozen quartile boundaries. G1 contains the smallest 25\% of targets, while G4 contains the
largest 25\%.}
    \label{tab:supp_target_area_groups}

    \textbf{OS-Atlas}\\[2pt]
    \begin{tabular}{lcccc}
        \toprule
       Objective & G1 & G2 & G3 & G4 \\
        \midrule
        MissClick-T
        & 87/186; 46.8\%
        & 88/185; 47.6\%
        & 78/199; 39.2\%
        & 92/199; 46.2\% \\
        Digit CE
        & 76/186; 40.9\%
        & 71/185; 38.4\%
        & 58/199; 29.1\%
        & 66/199; 33.2\% \\
        Soft-coordinate distance
        & 34/186; 18.3\%
        & 35/185; 18.9\%
        & 29/199; 14.6\%
        & 23/199; 11.6\% \\
        Token~CE
        & 36/186; 19.4\%
        & 27/185; 14.6\%
        & 20/199; 10.1\%
        & 18/199; 9.0\% \\
        \bottomrule
    \end{tabular}

    \medskip
    \textbf{UGround}\\[2pt]
    \begin{tabular}{lcccc}
        \toprule
       Objective & G1 & G2 & G3 & G4 \\
        \midrule
        MissClick-T
        & 115/193; 59.6\%
        & 123/202; 60.9\%
        & 127/201; 63.2\%
        & 137/205; 66.8\% \\
        Digit CE
        & 95/193; 49.2\%
        & 104/202; 51.5\%
        & 105/201; 52.2\%
        & 102/205; 49.8\% \\
        Soft-coordinate distance
        & 49/193; 25.4\%
        & 59/202; 29.2\%
        & 55/201; 27.4\%
        & 58/205; 28.3\% \\
        Token~CE
        & 34/193; 17.6\%
        & 39/202; 19.3\%
        & 25/201; 12.4\%
        & 27/205; 13.2\% \\
        \bottomrule
    \end{tabular}
\end{table*}

\begin{table*}[t]
    \centering
    \small
    \setlength{\tabcolsep}{3.5pt}
    \caption{Targeted ASR by fixed source-to-target
    center-distance quartile. Each cell reports successes/tasks
    and ASR. The two model panels use the same frozen quartile
    boundaries. G1 contains the 25\% of tasks with the smallest source-to-target center distances, while G4 contains the 25\% with the largest distances.}
    \label{tab:supp_target_distance_groups}

    \textbf{OS-Atlas}\\[2pt]
    \begin{tabular}{lcccc}
        \toprule
        Objective & G1 & G2 & G3 & G4 \\
        \midrule
        MissClick-T
        & 130/189; 68.8\%
        & 94/193; 48.7\%
        & 59/193; 30.6\%
        & 62/194; 32.0\% \\
        Digit CE
        & 97/189; 51.3\%
        & 78/193; 40.4\%
        & 47/193; 24.4\%
        & 49/194; 25.3\% \\
        Soft-coordinate distance
        & 51/189; 27.0\%
        & 28/193; 14.5\%
        & 17/193; 8.8\%
        & 25/194; 12.9\% \\
        Token~CE
        & 38/189; 20.1\%
        & 21/193; 10.9\%
        & 13/193; 6.7\%
        & 29/194; 14.9\% \\
        \bottomrule
    \end{tabular}

    \medskip
    \textbf{UGround}\\[2pt]
    \begin{tabular}{lcccc}
        \toprule
        Objective & G1 & G2 & G3 & G4 \\
        \midrule
        MissClick-T
        & 155/200; 77.5\%
        & 133/200; 66.5\%
        & 110/201; 54.7\%
        & 104/200; 52.0\% \\
        Digit CE
        & 125/200; 62.5\%
        & 105/200; 52.5\%
        & 89/201; 44.3\%
        & 87/200; 43.5\% \\
        Soft-coordinate distance
        & 63/200; 31.5\%
        & 63/200; 31.5\%
        & 48/201; 23.9\%
        & 47/200; 23.5\% \\
        Token~CE
        & 30/200; 15.0\%
        & 34/200; 17.0\%
        & 24/201; 11.9\%
        & 37/200; 18.5\% \\
        \bottomrule
    \end{tabular}
\end{table*}

\subsection{Target-Region Construction}
\label{sec:supp_target_construction}
\paragraph{Candidate Selection and Sampling.}
\label{par:supp_target_sampling}
Target regions are constructed before applying model-specific
clean-correct filtering. For each ScreenSpot-v2 task, we consider
the annotated UI elements in the same screenshot as candidate
targets. A valid candidate must correspond to an element
distinct from the ground-truth element and must have zero
intersection-over-union with $B_{\mathrm{gt}}$.
If at least one valid candidate exists, one candidate is selected
uniformly at random. Tasks without a valid
candidate are excluded from targeted evaluation. This procedure
produces a common target-construction set of $896$ tasks:
$178$ desktop, $323$ web, and $395$ mobile tasks.
The selected source--target pair is fixed before attack
generation and shared across victim models and targeted methods.
\paragraph{Model-Specific Target References.}
\label{par:supp_target_references}
The selected target element defines a common target region
$B_{\mathrm{tgt}}$ in the normalized screen coordinate system.
For OS-Atlas, the teacher-forced target reference contains the
target bounding-box coordinates in the native OS-Atlas answer
format. For UGround, the target reference contains the
model-native point corresponding to the target-element center.
These reference formats affect only the teacher-forced
optimization sequence. For both victim models, targeted attack
success is evaluated using the same complete target region:
the final freely generated executed click must fall inside
$B_{\mathrm{tgt}}$. The attack is not required to exactly
reproduce the reference coordinate string.

\subsection{Target-Region Statistics}
\label{sec:supp_target_statistics}

The geometry statistics are computed from the normalized
$[0,1]$ coordinates used in the evaluation JSON files. For each
source--target pair, target area is the area of
$B_{\mathrm{tgt}}$, and source-to-target center distance is the
Euclidean distance between the centers of $B_{\mathrm{gt}}$ and
$B_{\mathrm{tgt}}$.

Table~\ref{tab:supp_target_statistics} reports these quantities
and source--target IoU for the common 896-task set, both overall
and by platform. Target regions differ in size and in their
distance from the original element. For all 896 tasks, the source
and target bounding boxes have zero IoU.

\section{Targeted Attack Results}

\subsection{Results by Element Type}
\label{sec:supp_element_type}
ScreenSpot-v2 categorizes each task element as either text or
icon/widget. Table~\ref{tab:supp_element_type} reports targeted
ASR separately for these two element types.
MissClick-T achieves the highest targeted ASR in all four
model--element-type groups. Compared with the strongest
alternative, Digit CE, MissClick-T improves ASR by $8.7$ and
$11.0$ percentage points on OS-Atlas text and icon/widget tasks,
and by $12.2$ and $11.6$ percentage points on the corresponding
UGround tasks. These results show that the advantage of
MissClick-T is observed for both element types and both victim
models.

\begin{figure}[t]
    \centering
    \begin{subfigure}[t]{0.48\linewidth}
        \centering
        \includegraphics[width=\linewidth]
        {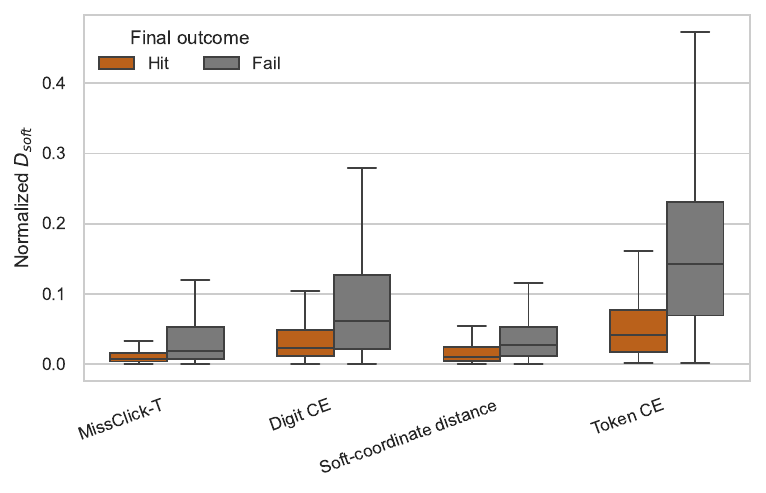}
        \caption{OS-Atlas: soft distance.}
    \end{subfigure}\hfill
    \begin{subfigure}[t]{0.48\linewidth}
        \centering
        \includegraphics[width=\linewidth]
        {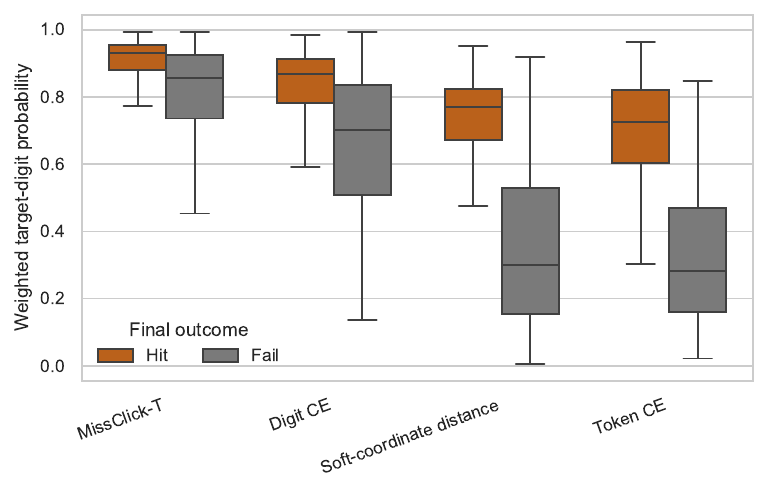}
        \caption{OS-Atlas: target probability.}
    \end{subfigure}

    \begin{subfigure}[t]{0.48\linewidth}
        \centering
        \includegraphics[width=\linewidth]
        {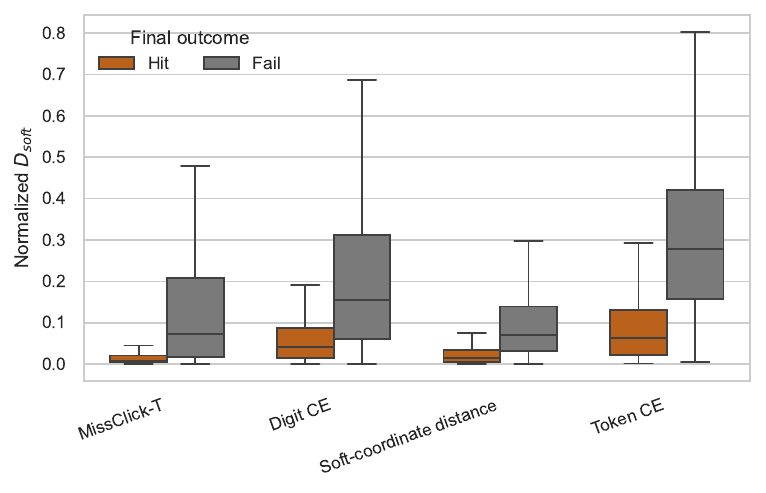}
        \caption{UGround: soft distance.}
    \end{subfigure}\hfill
    \begin{subfigure}[t]{0.48\linewidth}
        \centering
        \includegraphics[width=\linewidth]
        {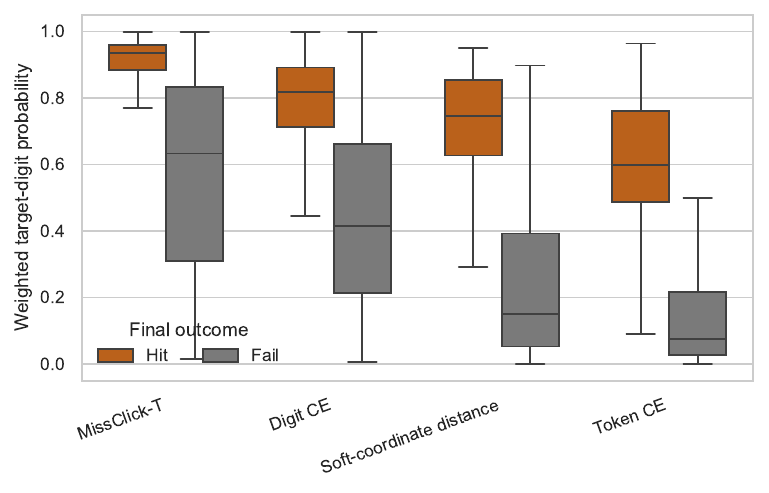}
        \caption{UGround: target probability.}
    \end{subfigure}
    \caption{Teacher-forced quantities stratified by final
    targeted outcome. Top row: OS-Atlas. Bottom row: UGround. Left column:
normalized soft target distance $D_{\mathrm{soft}}$. Right
column: place-weighted target-digit probability
$P_{\mathrm{tgt}}^{\mathrm{w}}$.}
    \label{fig:supp_boxplot}
\end{figure}

\subsection{Target-Region Size and Distance Analysis}
We analyze target size and source-to-target distance to determine
whether the advantage of MissClick-T is limited to large or nearby
targets. We define four target-area groups and four center-distance groups
using the common 896-task set. We use quartiles to obtain four
similarly sized groups without manually choosing area or distance
thresholds. All models and targeted objectives use the same group
boundaries. The target-area boundaries are $0.353\%$, $0.798\%$,
and $2.173\%$, and the normalized center-distance boundaries are
$0.254889$, $0.422449$, and $0.634902$.

Tables~\ref{tab:supp_target_area_groups} and
\ref{tab:supp_target_distance_groups} report the number of tasks
and targeted ASR in each group.
MissClick-T achieves the highest targeted ASR in every target-area
and center-distance group for both models. Targeted ASR generally
decreases as the source-to-target distance increases, but
MissClick-T remains the strongest method even in the farthest
group, exceeding Digit CE by $6.7$ percentage points on OS-Atlas
and $8.5$ percentage points on UGround. The target-area patterns
differ between the two models and do not show a common monotonic
trend. Overall, the advantage of MissClick-T is not limited to
large or nearby targets.

\section{Analysis of Targeted Optimization and Generation}
\label{sec:supp_analysis}

MissClick optimizes a teacher-forced objective but evaluates
on freely generated outputs. This section examines whether the
teacher-forced optimization signals are consistent with the
final targeted outcomes, and how place-value weighting changes
digit-level target alignment relative to uniform digit
weighting.
\subsection{Optimization Signal and Final Outcome}
\label{sec:supp_hit_fail}
For each targeted sample, we record two quantities from the
teacher-forced forward pass after optimization: the normalized
soft target distance $D_{\mathrm{soft}}$ (how close the
soft-decoded coordinate is to the target center) and the
place-weighted target-digit probability
$P_{\mathrm{tgt}}^{\mathrm{w}}$ (how confidently the model
assigns probability to the correct target digits, weighted by
decimal place value). We then run greedy autoregressive
generation on the same adversarial image and record the
normalized hard distance $D_{\mathrm{hard}}$ and whether the
executed click falls inside the target region (Hit or Fail).

Table~\ref{tab:supp_hit_fail_diagnostic} reports these
quantities stratified by final outcome.
Figure~\ref{fig:supp_boxplot} shows the corresponding
distributions. Across all four objectives and both models,
Hit samples have lower $D_{\mathrm{soft}}$ and higher
$P_{\mathrm{tgt}}^{\mathrm{w}}$ than Fail samples. For
MissClick-T, Hit samples have a median $D_{\mathrm{soft}}$
of $0.008$ on both models, compared to $0.019$ and $0.072$
for Fail samples on OS-Atlas and UGround, respectively.
The corresponding median $P_{\mathrm{tgt}}^{\mathrm{w}}$
is $0.931$ for Hit and $0.856$ for Fail samples on
OS-Atlas, and $0.935$ for Hit and $0.634$ for Fail samples
on UGround.

Among Fail samples, Soft-coordinate distance has a relatively
small $D_{\mathrm{soft}}$ but substantially lower
$P_{\mathrm{tgt}}^{\mathrm{w}}$ and $R_{\mathrm{argmax}}$
than MissClick-T and Digit CE. This pattern suggests that a small teacher-forced
soft-coordinate distance can coexist with weak target-digit
alignment and does not ensure successful final autoregressive
generation.

\begin{figure*}[t]
    \centering
 \begin{subfigure}[t]{0.235\textwidth}\centering
        \includegraphics[width=\linewidth]{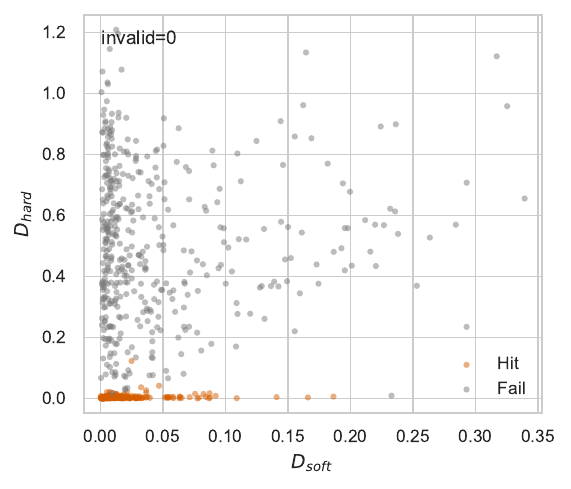}
        \caption{OS-Atlas, MissClick-T.}
    \end{subfigure}\hfill
    \begin{subfigure}[t]{0.235\textwidth}\centering
        \includegraphics[width=\linewidth]{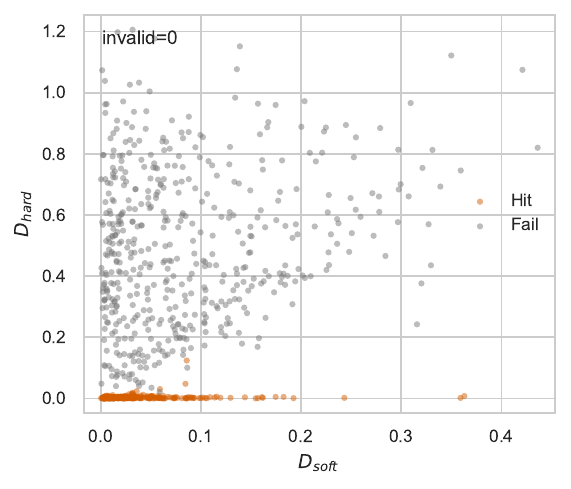}
        \caption{OS-Atlas, Digit CE.}
    \end{subfigure}\hfill
    \begin{subfigure}[t]{0.235\textwidth}\centering
        \includegraphics[width=\linewidth]{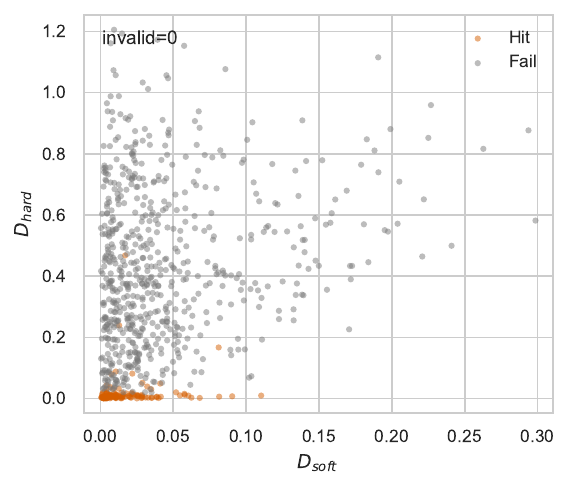}
        \caption{OS-Atlas, soft-coordinate distance.}
    \end{subfigure}\hfill
    \begin{subfigure}[t]{0.235\textwidth}\centering
        \includegraphics[width=\linewidth]{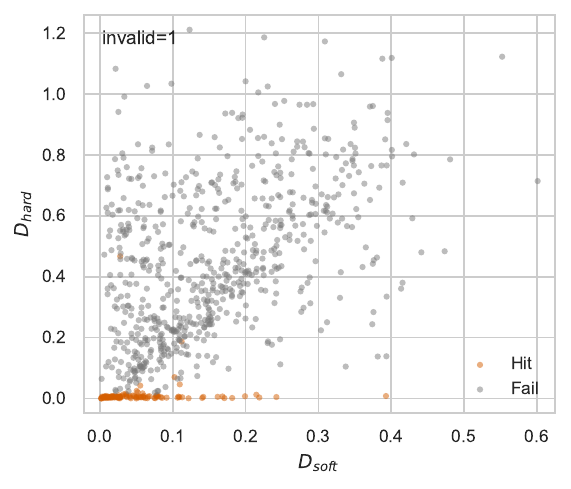}
        \caption{OS-Atlas, Token CE.}
    \end{subfigure}

    \begin{subfigure}[t]{0.235\textwidth}\centering
        \includegraphics[width=\linewidth]{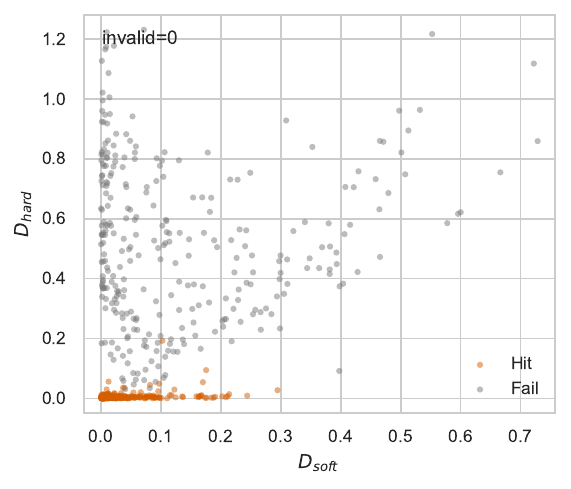}
        \caption{UGround, MissClick-T.}
    \end{subfigure}\hfill
    \begin{subfigure}[t]{0.235\textwidth}\centering
        \includegraphics[width=\linewidth]{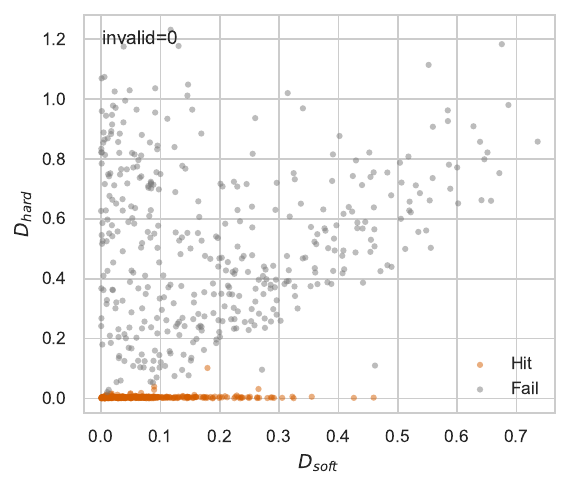}
        \caption{UGround, Digit CE.}
    \end{subfigure}\hfill
    \begin{subfigure}[t]{0.235\textwidth}\centering
        \includegraphics[width=\linewidth]{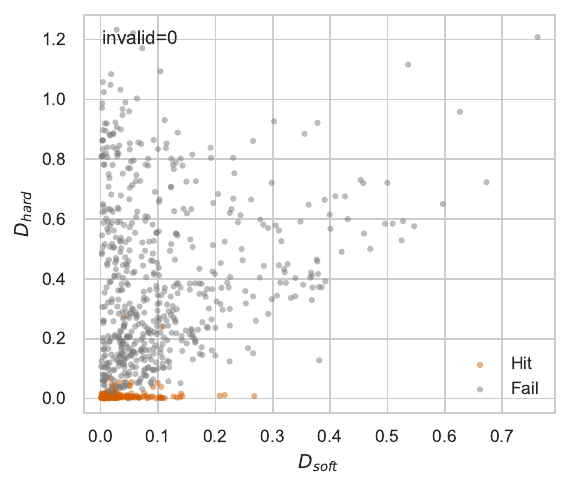}
        \caption{UGround, soft-coordinate distance.}
    \end{subfigure}\hfill
    \begin{subfigure}[t]{0.235\textwidth}\centering
        \includegraphics[width=\linewidth]{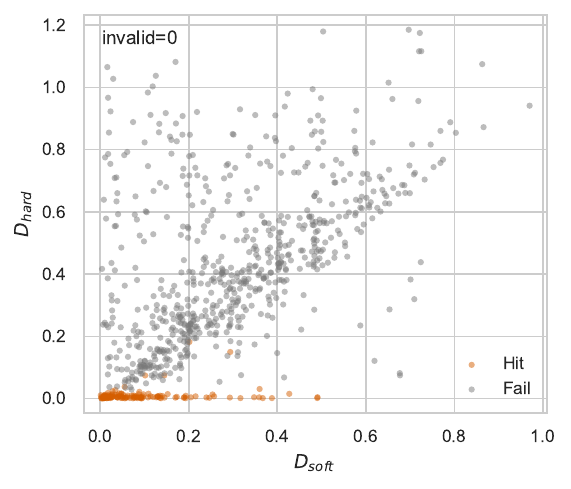}
        \caption{UGround, Token CE.}
    \end{subfigure}
    \caption{Soft target distance $D_{\mathrm{soft}}$ versus
    final greedy-generated target distance $D_{\mathrm{hard}}$,
    colored by final outcome (Hit or Fail). Each panel shows one
    model--objective combination.}
    \vspace{-0.5em}
    \label{fig:supp_scatter}
\end{figure*}

Figure~\ref{fig:supp_scatter} plots $D_{\mathrm{soft}}$
against $D_{\mathrm{hard}}$ for each sample, colored by
final outcome. Hit samples cluster in the lower-left corner
(small $D_{\mathrm{soft}}$, small $D_{\mathrm{hard}}$),
while Fail samples are spread toward higher values on both
axes. This pattern is consistent across models and
objectives: samples with smaller soft target distance after
optimization tend to have smaller hard distance after
generation. The single invalid OS-Atlas Token~CE output
(Table~\ref{tab:supp_parse_validity}) remains in the Fail
count but is excluded from the scatter plot because it does
not have a parsed final coordinate.

These results indicate that the teacher-forced optimization
signal and the final greedy-generated outcome maintain a
consistent relationship. They do not establish that a
successful optimization signal guarantees a successful final
outcome, as the final result is determined solely by the
greedy-generated click.

\subsection{Effect of Place-Value Weighting on Digit Positions}
\label{sec:supp_digit_position}

MissClick-T weights the cross-entropy at each digit position
by its decimal place value, while Digit CE assigns equal
weight to all positions. Table~\ref{tab:supp_digit_position}
compares the per-position results of the two objectives,
grouped by place value (hundreds, tens, ones).

On both models, MissClick-T achieves higher mean target
probability and argmax accuracy at the hundreds and tens
positions than Digit CE. On OS-Atlas, the hundreds-place
target probability is $0.922$ for MissClick-T versus $0.818$
for Digit CE; the tens-place values are $0.730$ versus
$0.671$. UGround shows the same pattern: $0.851$ versus
$0.691$ at the hundreds place, $0.710$ versus $0.612$ at the
tens place.

\begin{figure}[ht]
\vspace{-0.3em}
    \centering
    \begin{subfigure}[t]{0.48\linewidth}
        \centering
        \includegraphics[width=\linewidth]
        {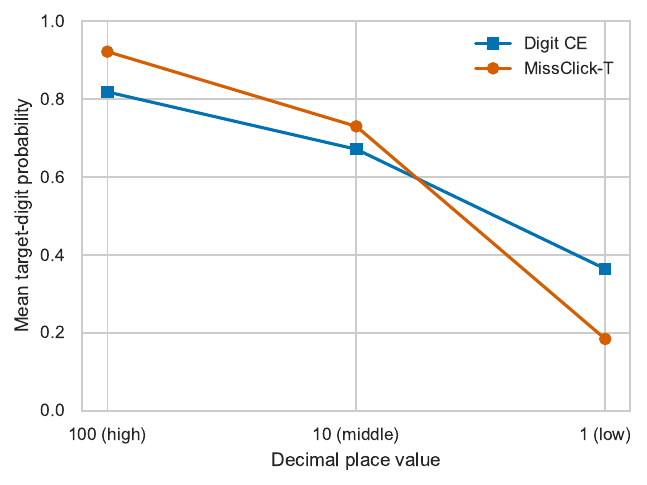}
        \caption{OS-Atlas}
    \end{subfigure}\hfill
    \begin{subfigure}[t]{0.48\linewidth}
        \centering
        \includegraphics[width=\linewidth]
        {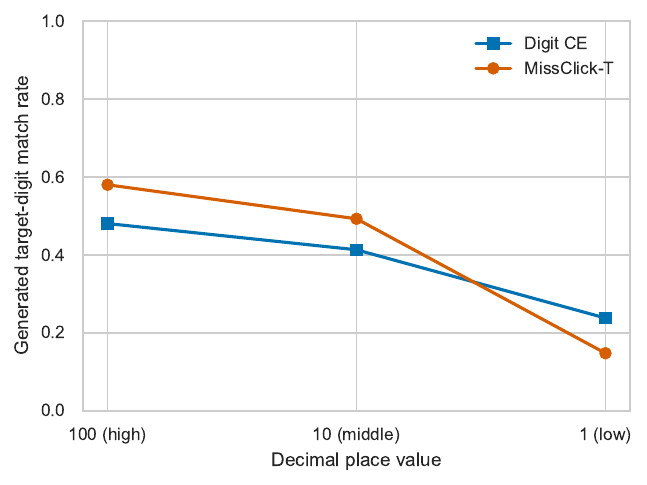}
        \caption{OS-Atlas}
    \end{subfigure}

    \begin{subfigure}[t]{0.48\linewidth}
        \centering
        \includegraphics[width=\linewidth]
        {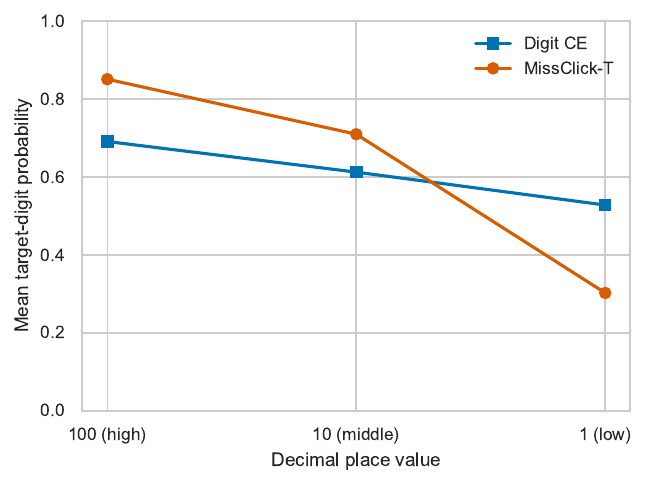}
        \caption{UGround}
    \end{subfigure}\hfill
    \begin{subfigure}[t]{0.48\linewidth}
        \centering
        \includegraphics[width=\linewidth]
        {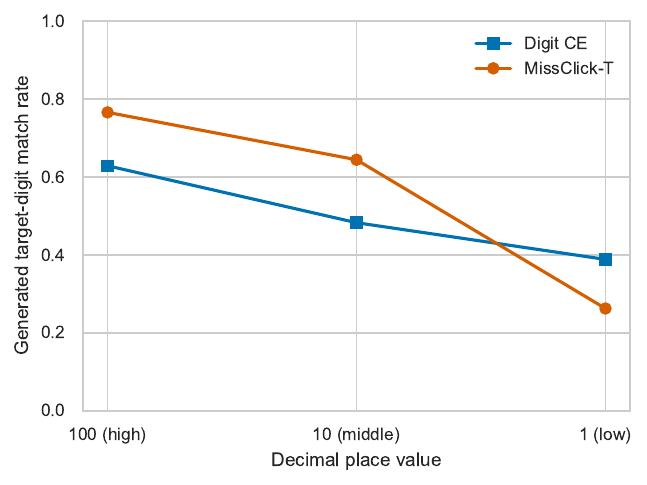}
        \caption{UGround}
    \end{subfigure}
    \vspace{-0.5em}
    \caption{Per-position comparison of uniform Digit CE and
    place-weighted MissClick-T. 
    Left column: mean target-digit
    probability. Right column: generated target-digit match rate.
    Top row: OS-Atlas. Bottom row: UGround.
    }
    \label{fig:supp_digit_place}
    \vspace{-1em}
\end{figure}

At the ones place, the pattern reverses: MissClick-T has
lower target probability than Digit CE on both models
($0.184$ vs.\ $0.364$ on OS-Atlas; $0.302$ vs.\ $0.528$ on
UGround). This trade-off is consistent with the method
design. A change in the hundreds digit shifts the coordinate
by approximately $100$ coordinate units in the $[0,1000]$
system, whereas a change in the ones digit shifts it by
approximately $1$ coordinate unit. Place-value weighting
therefore gives greater priority to digits with a larger
numerical effect on the generated coordinate. The observed
redistribution toward the hundreds and tens positions is
consistent with the higher targeted ASR of MissClick-T over
Digit CE in the main results.
Figure~\ref{fig:supp_digit_place} visualizes this comparison.
MissClick-T achieves higher target probability and match rate
at the hundreds and tens positions, while Digit CE is higher
at the ones position. This pattern is consistent across both
models and both metrics.

\section{Transferability and Input-Preprocessing Robustness}
\label{sec:supp_transfer_defense}

\subsection{Cross-Model Transferability}
\label{sec:supp_transfer}

To evaluate cross-model transferability, we use the adversarial
images generated against one victim model and evaluate them on
the other model using the same evaluation protocol. No
re-optimization is performed on the target model.

Table~\ref{tab:supp_transfer} reports untargeted and targeted
transfer ASR in both directions. In the untargeted setting,
transfer from UGround to OS-Atlas is substantially higher than
from OS-Atlas to UGround for both methods. MissClick-U achieves
higher untargeted transfer than Representation attack in the
OS-Atlas$\,{\to}\,$UGround direction (17.01\% vs.\ 14.16\%),
while the reverse direction favors Representation attack
(54.72\% vs.\ 38.56\%). The transfer results therefore vary
with both the transfer direction and the attack objective.

In the targeted setting, transfer rates are low for both methods
and both directions. MissClick-T nevertheless achieves roughly
twice the targeted transfer rate of Token~CE in both directions
(3.37\% vs.\ 1.75\% for OS-Atlas$\,{\to}\,$UGround; 8.71\%
vs.\ 4.16\% for UGround$\,{\to}\,$OS-Atlas).

Overall, cross-model transfer is substantially weaker than
white-box attack performance and varies by transfer direction
and attack objective.

\subsection{Robustness to Input Preprocessing}
\label{sec:supp_defense}

To examine whether simple non-adaptive input preprocessing
reduces attack effectiveness, we apply JPEG compression
(quality$=75$) and Gaussian blur ($\sigma{=}1.0$, kernel
size$=3$) to both clean and adversarial images before
evaluation. These defenses are non-adaptive: the attack is
not aware of the preprocessing at optimization time.

Table~\ref{tab:supp_defense} reports clean accuracy and the
ASR of MissClick-U and MissClick-T under each preprocessing.
Both defenses reduce clean accuracy. JPEG-75 lowers OS-Atlas
clean accuracy from 84.20\% to 79.25\% and UGround from
88.29\% to 87.50\%; Gaussian blur lowers them to 80.82\% and
87.66\%, respectively.

Both defenses also reduce MissClick ASR substantially.
Untargeted ASR drops from 75.07\% to below 21\% on OS-Atlas
and from 72.93\% to below 17\% on UGround. Targeted ASR is
reduced even more sharply, falling below 5\% for both models
under either defense.

These results apply only to the evaluated non-adaptive
setting and do not establish robustness against an adaptive
attack optimized through the preprocessing operation.
\begin{table*}[t]
    \centering
    \small
    \setlength{\tabcolsep}{3.5pt}
    \caption{Hit/Fail diagnostic. Values are median (IQR); $n$ is the final-outcome task count, $n_d$ is the number with a valid teacher-forced diagnostic, and $n_h$ is the number with a valid final hard distance.}
    \label{tab:supp_hit_fail_diagnostic}
    \resizebox{\linewidth}{!}{%
    \begin{tabular}{lllrrrrrr}
        \toprule
        Model & Objective & Outcome & $n$ & $n_d$ & $D_{\mathrm{soft}}$ & $P_{\mathrm{tgt}}^{\mathrm{w}}$ & $R_{\mathrm{argmax}}$ & $D_{\mathrm{hard}}$ / $D_{\mathrm{region}}$ ($n_h$) \\
        \midrule
        OS-Atlas & MissClick-T & Hit & 345 & 345 & 0.008 (0.012) & 0.931 (0.075) & 0.700 (0.114) & 0.003 (0.003) / 0.000 (0.000) (345) \\
         &  & Fail & 424 & 424 & 0.019 (0.046) & 0.856 (0.189) & 0.636 (0.182) & 0.494 (0.362) / 0.417 (0.368) (424) \\
         & Digit CE & Hit & 271 & 271 & 0.023 (0.038) & 0.869 (0.133) & 0.818 (0.217) & 0.002 (0.003) / 0.000 (0.000) (271) \\
         &  & Fail & 498 & 498 & 0.061 (0.105) & 0.702 (0.327) & 0.727 (0.230) & 0.473 (0.383) / 0.385 (0.360) (498) \\
         & \shortstack[l]{Soft-coordinate\\distance} & Hit & 121 & 121 & 0.011 (0.020) & 0.770 (0.151) & 0.500 (0.167) & 0.007 (0.010) / 0.000 (0.000) (121) \\
         &  & Fail & 648 & 648 & 0.028 (0.042) & 0.300 (0.376) & 0.333 (0.194) & 0.396 (0.372) / 0.318 (0.375) (648) \\
         & Token CE & Hit & 101 & 101 & 0.041 (0.059) & 0.726 (0.218) & 0.600 (0.167) & 0.005 (0.004) / 0.000 (0.000) (101) \\
         &  & Fail & 668 & 668 & 0.143 (0.162) & 0.283 (0.312) & 0.455 (0.220) & 0.430 (0.386) / 0.353 (0.378) (667) \\
        \midrule
        UGround & MissClick-T & Hit & 502 & 502 & 0.008 (0.017) & 0.935 (0.076) & 0.800 (0.167) & 0.004 (0.004) / 0.000 (0.000) (502) \\
         &  & Fail & 299 & 299 & 0.072 (0.192) & 0.634 (0.523) & 0.600 (0.167) & 0.487 (0.416) / 0.424 (0.410) (299) \\
         & Digit CE & Hit & 406 & 406 & 0.041 (0.074) & 0.819 (0.179) & 0.833 (0.200) & 0.004 (0.002) / 0.000 (0.000) (406) \\
         &  & Fail & 395 & 395 & 0.155 (0.251) & 0.414 (0.448) & 0.667 (0.333) & 0.502 (0.411) / 0.420 (0.419) (395) \\
         & \shortstack[l]{Soft-coordinate\\distance} & Hit & 221 & 221 & 0.014 (0.029) & 0.747 (0.228) & 0.667 (0.167) & 0.006 (0.006) / 0.000 (0.000) (221) \\
         &  & Fail & 580 & 580 & 0.071 (0.108) & 0.150 (0.340) & 0.333 (0.300) & 0.358 (0.405) / 0.292 (0.405) (580) \\
         & Token CE & Hit & 125 & 125 & 0.062 (0.109) & 0.598 (0.274) & 0.667 (0.200) & 0.005 (0.007) / 0.000 (0.000) (125) \\
         &  & Fail & 676 & 676 & 0.279 (0.262) & 0.075 (0.189) & 0.500 (0.267) & 0.406 (0.366) / 0.322 (0.363) (676) \\
        \bottomrule
    \end{tabular}
    }
\end{table*}

\begin{table*}[t]
    \centering
    \small
    \setlength{\tabcolsep}{3.5pt}
    \caption{Coordinate-digit results for uniform Digit CE and
    place-weighted MissClick-T. Each cell reports mean target
    probability, target-digit argmax accuracy, and generated
    target-digit match rate, in that order.}


    \label{tab:supp_digit_position}
    \begin{tabular}{llccc}
        \toprule
        Model & Objective
        & 100 (high) & 10 (middle) & 1 (low) \\
        \midrule
        OS-Atlas & Digit CE
        & 0.818 / 0.897 / 0.480
        & 0.671 / 0.808 / 0.413
        & 0.364 / 0.544 / 0.238 \\
        & MissClick-T
        & 0.922 / 0.961 / 0.580
        & 0.730 / 0.850 / 0.493
        & 0.184 / 0.235 / 0.147 \\
        \midrule
        UGround & Digit CE
        & 0.691 / 0.789 / 0.629
        & 0.612 / 0.709 / 0.483
        & 0.528 / 0.778 / 0.388 \\
        & MissClick-T
        & 0.851 / 0.900 / 0.766
        & 0.710 / 0.816 / 0.644
        & 0.302 / 0.426 / 0.262 \\
        \bottomrule
    \end{tabular}
\end{table*}

\begin{table*}[t]
    \centering
    \small
    \setlength{\tabcolsep}{3.5pt}
    \caption{Cross-model transfer attack success on ScreenSpot-v2
    ($\varepsilon{=}16/255$). Each reported rate is a micro-averaged ASR. Untargeted and targeted denominators
    are the target model's corresponding clean-correct sets
    pooled over desktop, web, and mobile.}
    \label{tab:supp_transfer}
    \begin{tabular}{llclc}
        \toprule
        Source $\rightarrow$ Target
        & Untargeted source attack & U-ASR
        & Targeted source attack & T-ASR \\
        \midrule
        OS-Atlas $\rightarrow$ UGround
        & Representation attack &  14.16\%
        & Token~CE &  1.75\% \\
        & MissClick-U & 17.01\%
        & MissClick-T &3.37\% \\
        \addlinespace
        UGround $\rightarrow$ OS-Atlas
        & Representation attack &  54.72\%
        & Token~CE &  4.16\% \\
        & MissClick-U & 38.56\%
        & MissClick-T & 8.71\% \\
        \bottomrule
    \end{tabular}
\end{table*}
\begin{table*}[t]
    \centering
    \small
    \setlength{\tabcolsep}{3.5pt}
    \caption{Effect of non-adaptive input preprocessing on
    clean accuracy and MissClick attack success
    ($\varepsilon{=}16/255$). Each cell reports the corresponding accuracy or attack success rate. Clean accuracy is evaluated on preprocessed original screenshots from the full ScreenSpot-v2 set, whereas U-ASR and T-ASR are evaluated on preprocessed adversarial screenshots generated from the frozen clean-correct task sets. All entries are micro-averages over desktop, web, and mobile.}
    \label{tab:supp_defense}
    \begin{tabular}{lcccccc}
        \toprule
        Defense & OS Clean & OS U-ASR & OS T-ASR
        & UG Clean & UG U-ASR & UG T-ASR \\
        \midrule
        None
        &  84.20\%
        &  75.07\%
        &  44.86\%
        &  88.29\%
        &  72.93\%
        &  62.67\% \\
        JPEG-75
        &  79.25\%
        &  19.61\%
        & 4.55\%
        &  87.50\%
        & 16.83\%
        & 3.75\% \\
        Gaussian blur
        &  80.82\%
        &20.92\%
        & 3.12\%
        & 87.66\%
        &  13.09\%
        & 2.25\% \\
        \bottomrule
    \end{tabular}
\end{table*}
\section{Attack Visualization}
\label{sec:supp_qualitative}

\providecommand{\attackpanelrow}[3]{%
\parbox{\textwidth}{\centering\small\textbf{#2}: ``#3''}\\[2pt]%
\includegraphics[width=0.325\textwidth]{#1_clean.png}\hfill%
\includegraphics[width=0.325\textwidth]{#1_adversarial.png}\hfill%
\includegraphics[width=0.325\textwidth]{#1_perturbation.png}\\[2pt]}

This section shows successful attack examples evaluated using
the same success criteria as the quantitative experiments.
The displayed clean and adversarial clicks are obtained from
the model outputs on the corresponding images.

\subsection{Untargeted Attack Examples}
\label{sec:supp_untargeted_examples}
\begin{figure*}[t]
    \centering

    \parbox{\textwidth}{
        \centering\small\textbf{Desktop}: ``sign in''
    }\\[2pt]
    \includegraphics[width=0.325\textwidth]
    {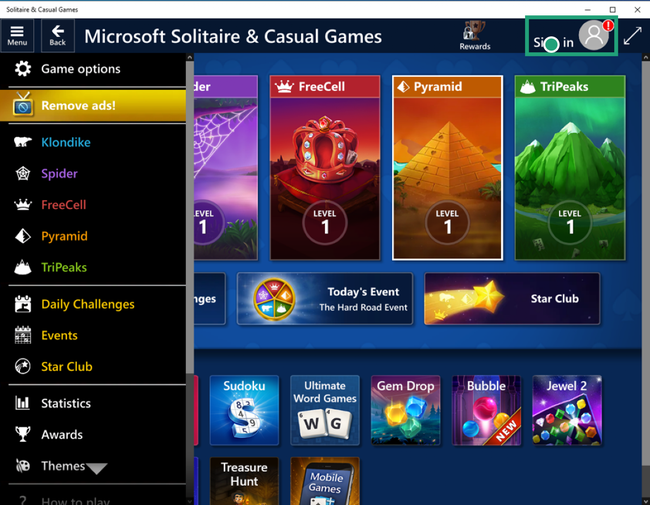}
    \hfill
    \includegraphics[width=0.325\textwidth]
    {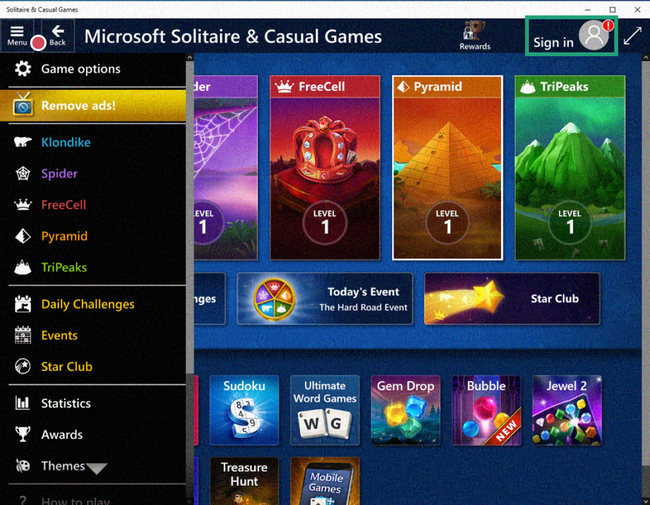}
    \hfill
    \includegraphics[width=0.325\textwidth]
    {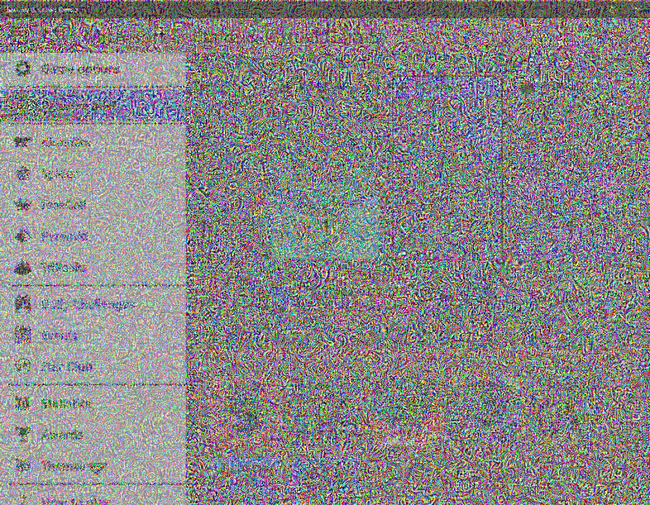}
    \\[2pt]

    \parbox{\textwidth}{
        \centering\small\textbf{Web}: ``close the cart''
    }\\[2pt]
    \includegraphics[width=0.325\textwidth]
    {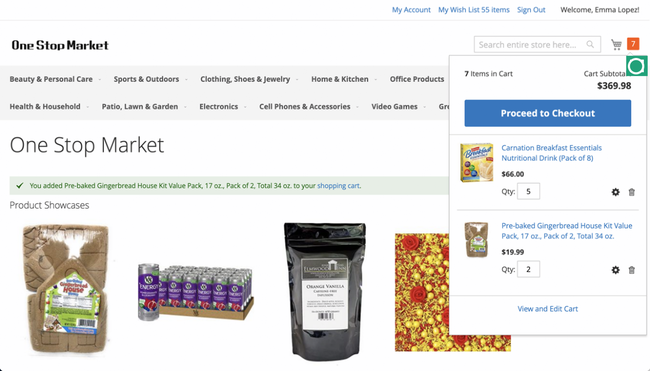}
    \hfill
    \includegraphics[width=0.325\textwidth]
    {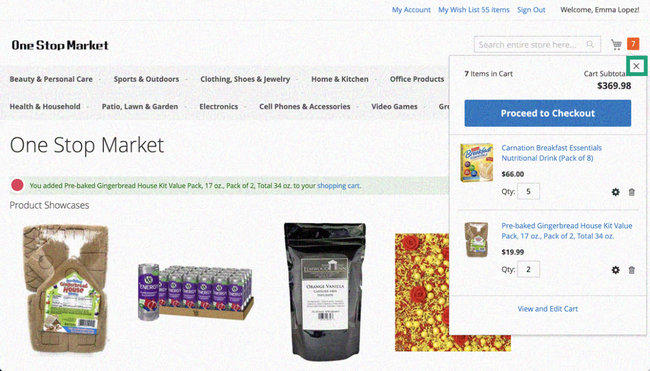}
    \hfill
    \includegraphics[width=0.325\textwidth]
    {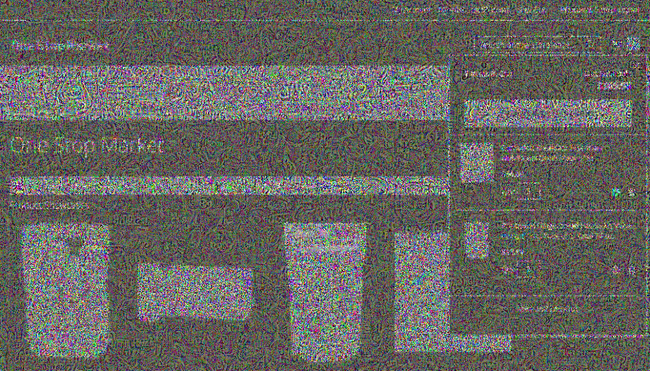}
    \\[2pt]

    \parbox{\textwidth}{
        \centering\small\textbf{Mobile}: ``add an event''
    }\\[2pt]
    \includegraphics[width=0.325\textwidth]
    {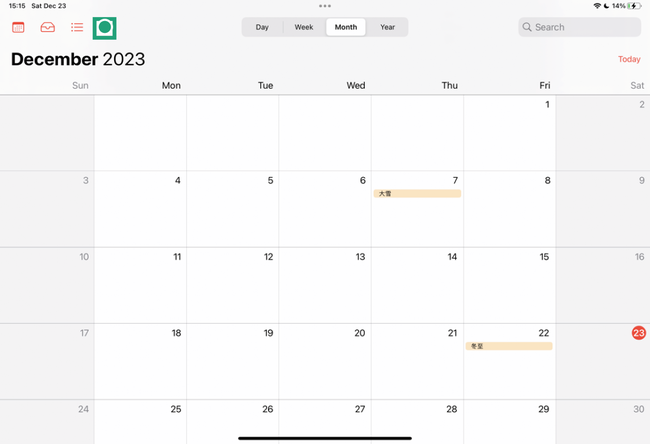}
    \hfill
    \includegraphics[width=0.325\textwidth]
    {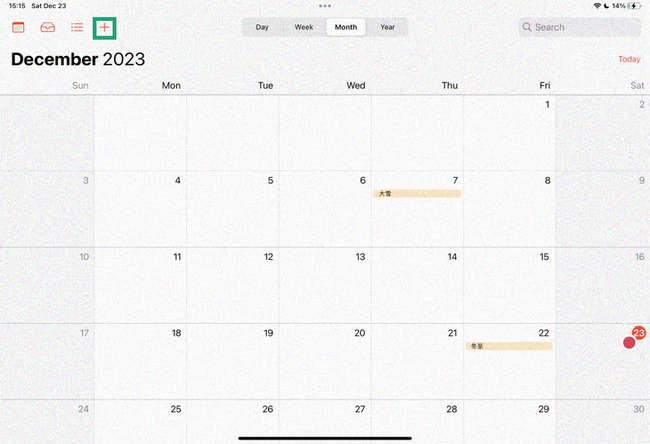}
    \hfill
    \includegraphics[width=0.325\textwidth]
    {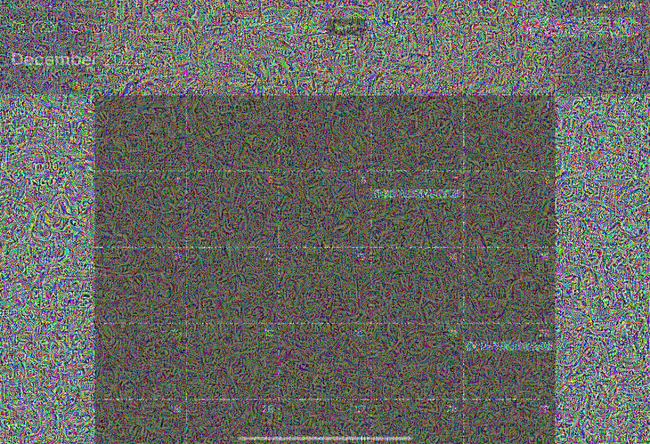}
    \\[2pt]

    \caption{Successful untargeted attack examples.}
    \label{fig:supp_untargeted_examples_os}
\end{figure*}

\begin{figure*}[t]
    \centering

    \parbox{\textwidth}{
        \centering\small\textbf{Desktop}: ``turn off the computer''
    }\\[2pt]
    \includegraphics[width=0.325\textwidth]
    {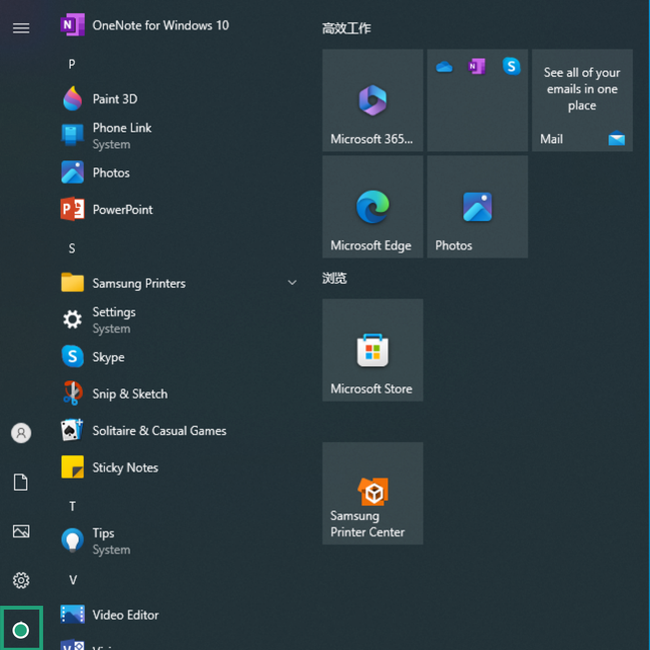}
    \hfill
    \includegraphics[width=0.325\textwidth]
    {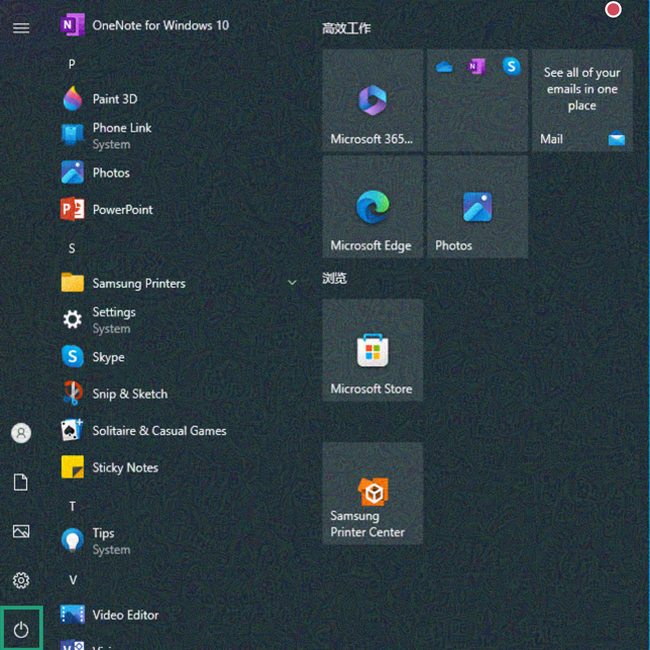}
    \hfill
    \includegraphics[width=0.325\textwidth]
    {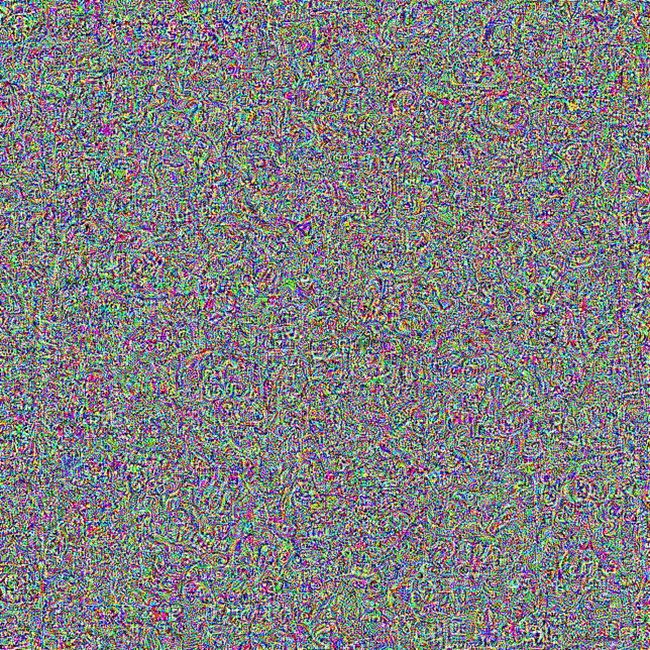}
    \\[2pt]

    \parbox{\textwidth}{
        \centering\small\textbf{Web}: ``print the page''
    }\\[2pt]
    \includegraphics[width=0.325\textwidth]
    {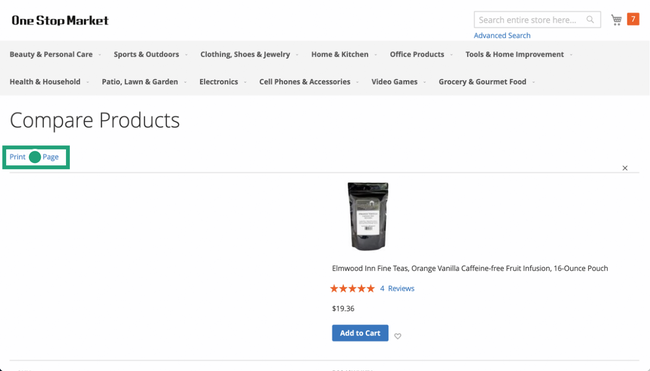}
    \hfill
    \includegraphics[width=0.325\textwidth]
    {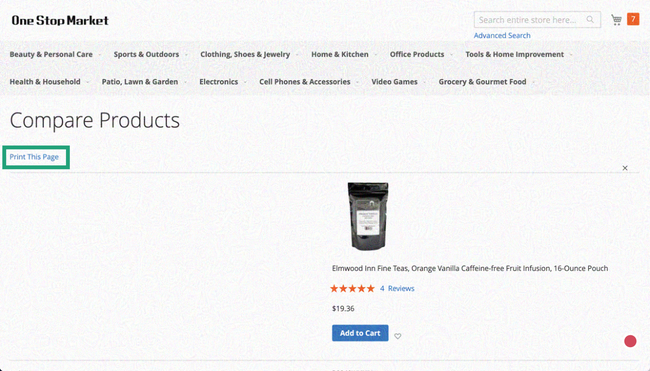}
    \hfill
    \includegraphics[width=0.325\textwidth]
    {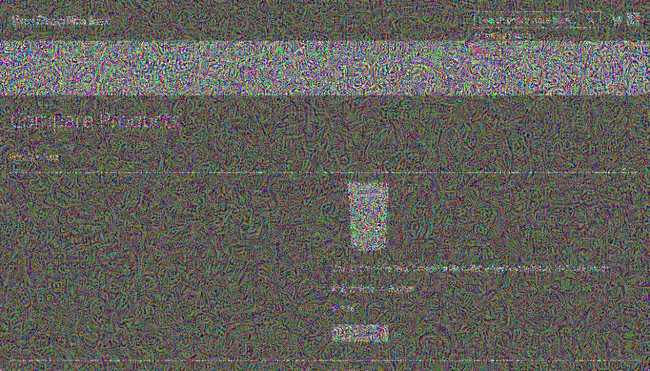}
    \\[2pt]

    \parbox{\textwidth}{
        \centering\small\textbf{Mobile}: ``check settings''
    }\\[2pt]
    \includegraphics[width=0.325\textwidth]
    {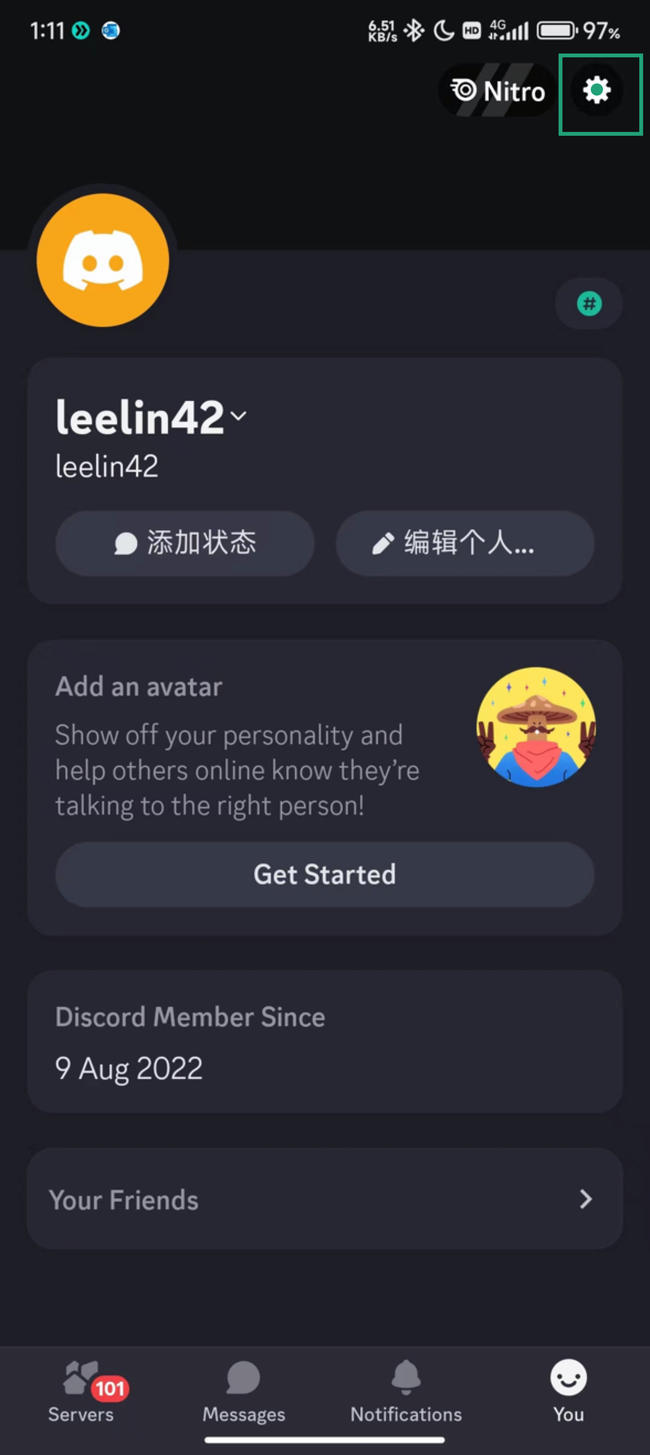}
    \hfill
    \includegraphics[width=0.325\textwidth]
    {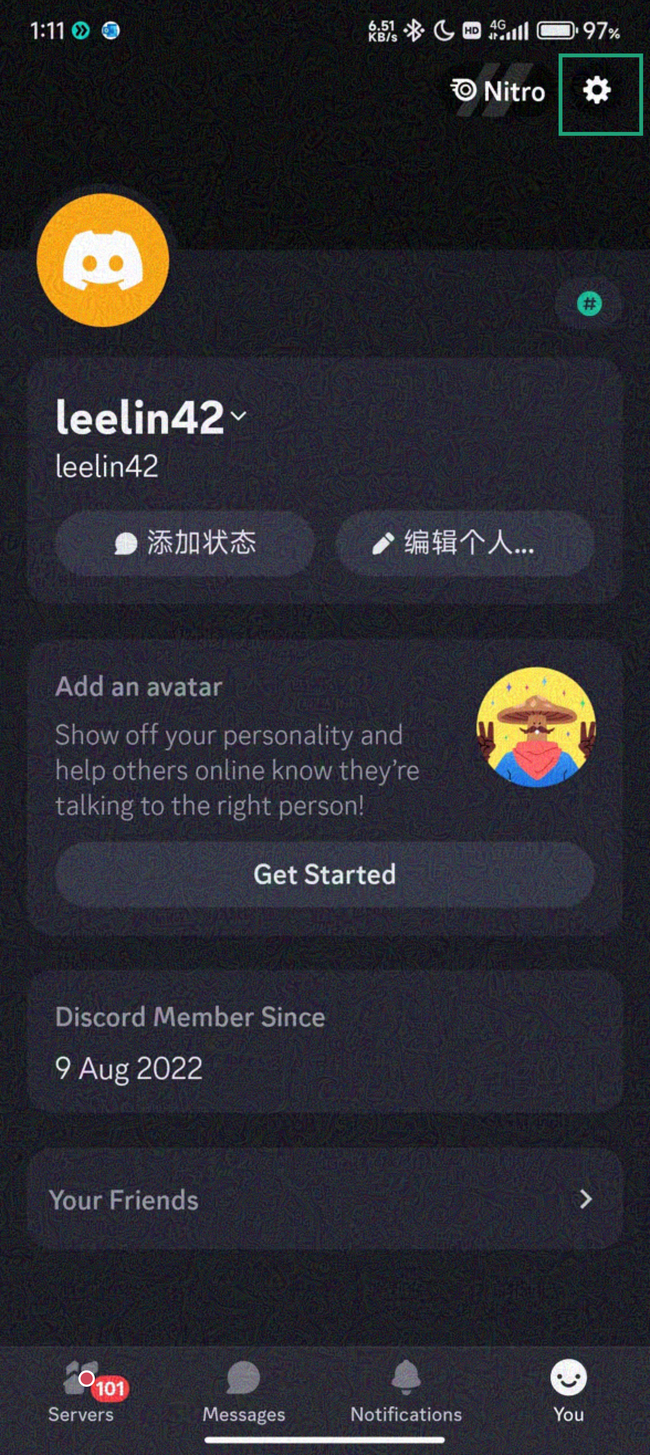}
    \hfill
    \includegraphics[width=0.325\textwidth]
    {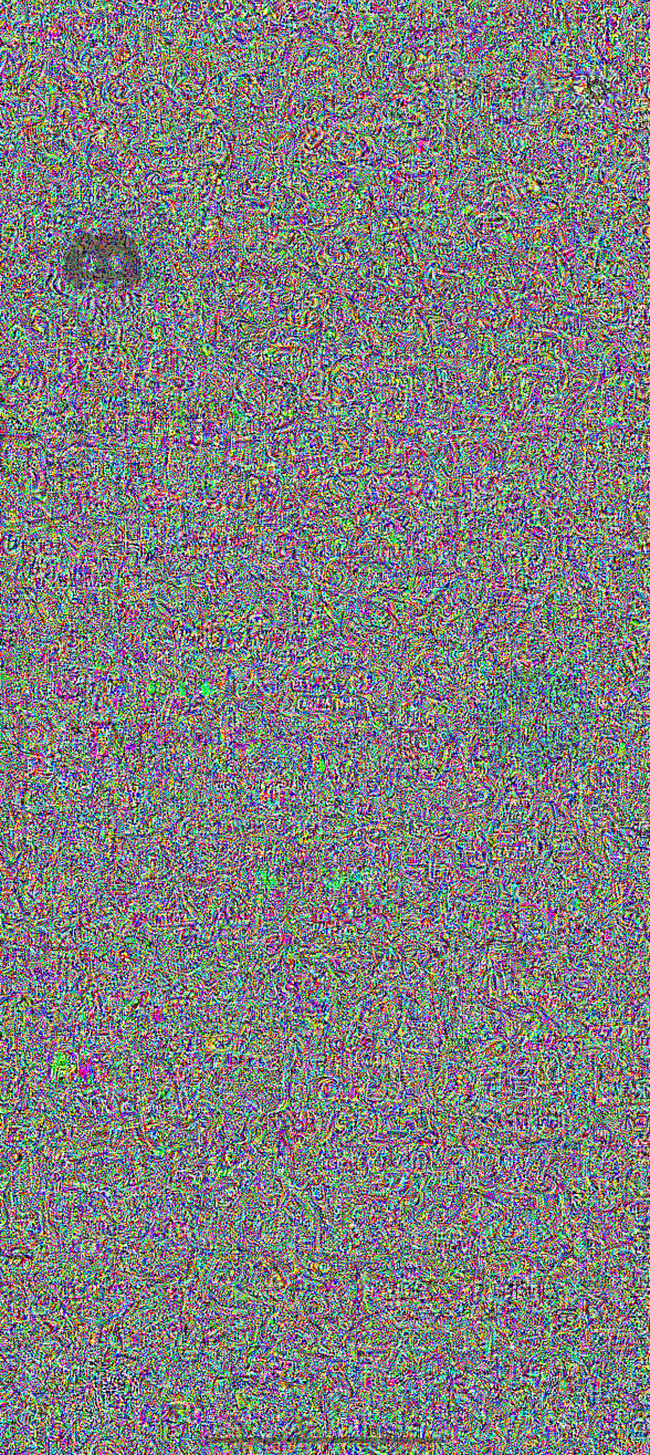}
    \\[2pt]

    \caption{Successful UGround MissClick-U examples at
    $\varepsilon=16/255$. The layout and annotation rules match
    Figure~\ref{fig:supp_untargeted_examples_os}.}
    \label{fig:supp_untargeted_examples_ug}
\end{figure*}

Figures~\ref{fig:supp_untargeted_examples_os} and
\ref{fig:supp_untargeted_examples_ug} show successful MissClick-U
examples from desktop, web, and mobile tasks. In each case, the
clean click lies inside the ground-truth region, whereas the
adversarial click lies outside it.

\subsection{Targeted Attack Examples}
\label{sec:supp_targeted_examples}
\begin{figure*}[t]
    \centering

    \parbox{\textwidth}{
        \centering\small\textbf{Desktop}: ``add a new account''
    }\\[2pt]
    \includegraphics[width=0.325\textwidth]
    {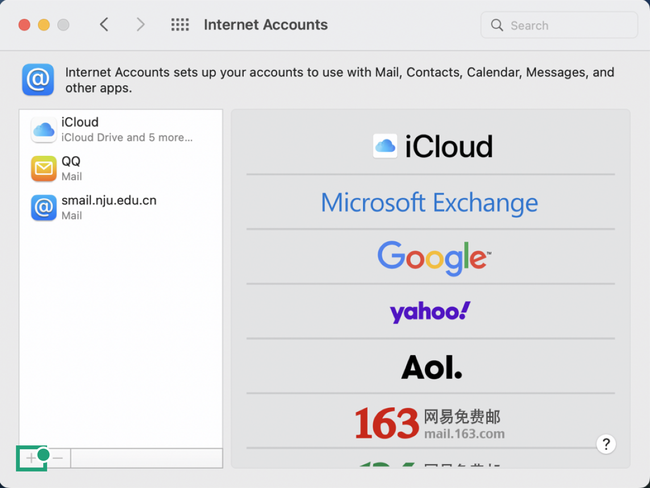}
    \hfill
    \includegraphics[width=0.325\textwidth]
    {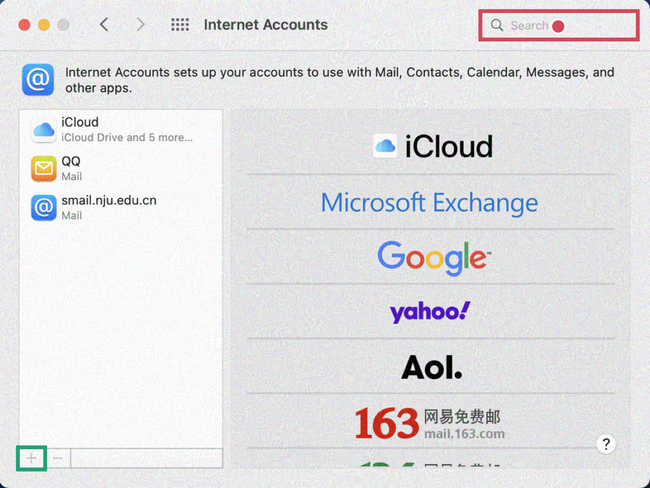}
    \hfill
    \includegraphics[width=0.325\textwidth]
    {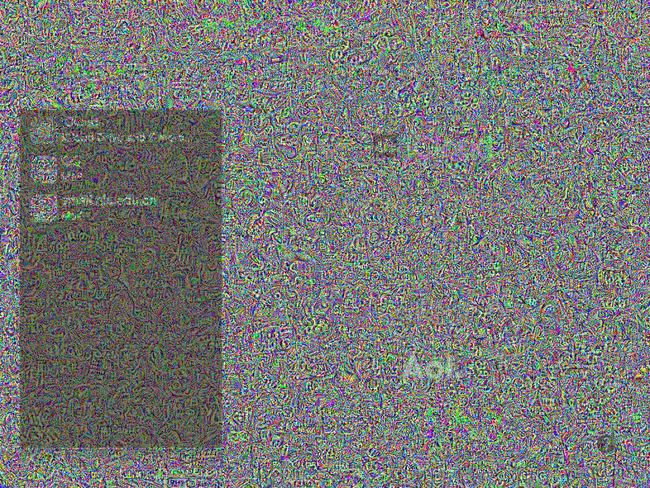}
    \\[2pt]

    \parbox{\textwidth}{
        \centering\small\textbf{Web}: ``change page language''
    }\\[2pt]
    \includegraphics[width=0.325\textwidth]
    {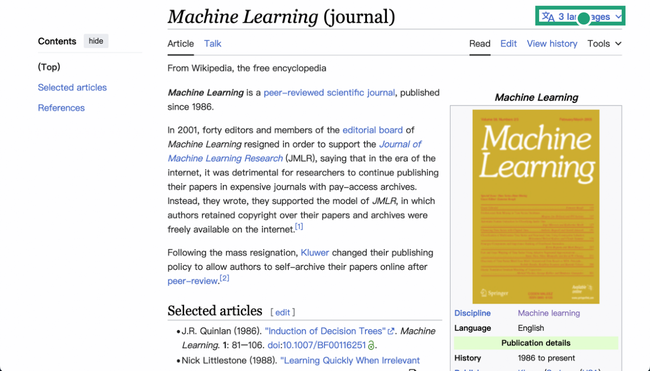}
    \hfill
    \includegraphics[width=0.325\textwidth]
    {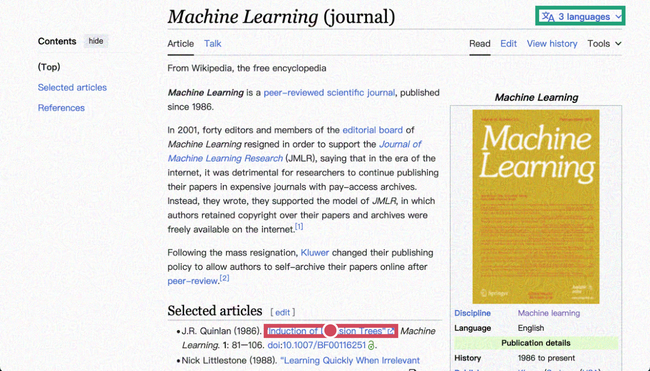}
    \hfill
    \includegraphics[width=0.325\textwidth]
    {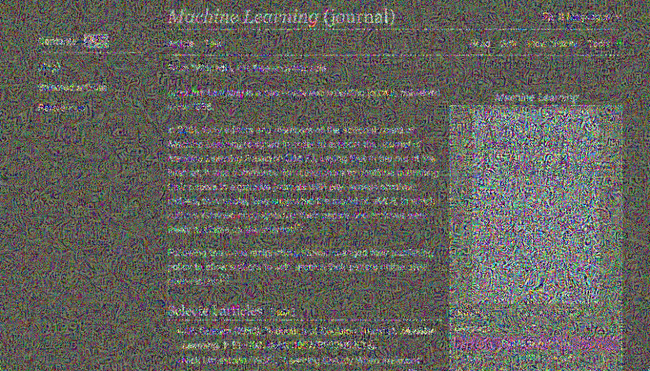}
    \\[2pt]

    \parbox{\textwidth}{
        \centering\small\textbf{Mobile}: ``add a new app to the desktop''
    }\\[2pt]
    \includegraphics[width=0.325\textwidth]
    {targeted_examples_os_atlas_os_mobile_mobile_263_eps16_clean.png}
    \hfill
    \includegraphics[width=0.325\textwidth]
    {targeted_examples_os_atlas_os_mobile_mobile_263_eps16_adversarial.png}
    \hfill
    \includegraphics[width=0.325\textwidth]
    {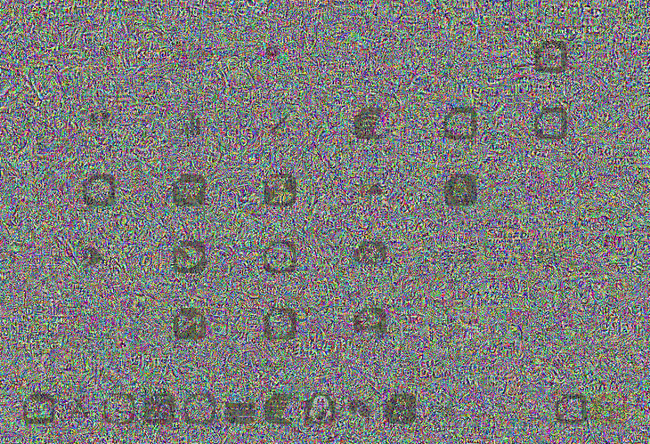}
    \\[2pt]

    \caption{Successful OS-Atlas MissClick-T examples at
    $\varepsilon=16/255$. Green boxes and points denote the
    ground-truth region and clean click. Red boxes denote the
    attacker-selected target region, and red points denote the
    final adversarial click. The third column shows the signed
    perturbation at a fixed $10\times$ scale.}
    \label{fig:supp_targeted_examples_os}
\end{figure*}

\begin{figure*}[t]
    \centering

    \parbox{\textwidth}{
        \centering\small\textbf{Desktop}: ``question about bluetooth''
    }\\[2pt]
    \includegraphics[width=0.325\textwidth]
    {targeted_examples_uground_ug_desktop_desktop_199_eps16_clean.png}
    \hfill
    \includegraphics[width=0.325\textwidth]
    {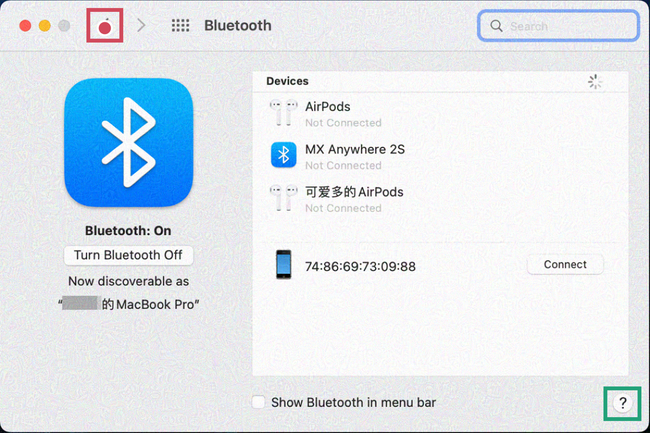}
    \hfill
    \includegraphics[width=0.325\textwidth]
    {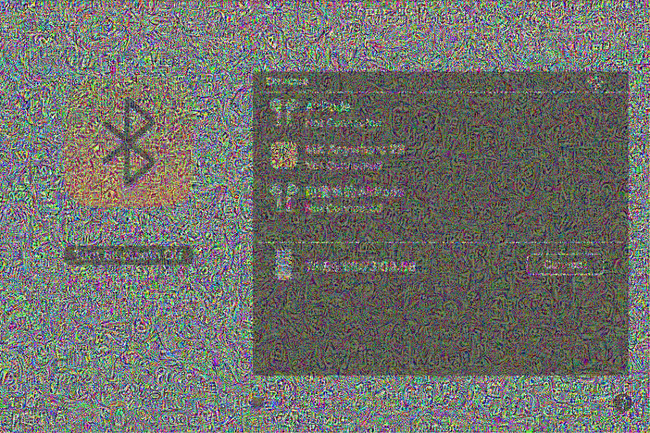}
    \\[2pt]

    \parbox{\textwidth}{
        \centering\small\textbf{Web}: ``change page language''
    }\\[2pt]
    \includegraphics[width=0.325\textwidth]
    {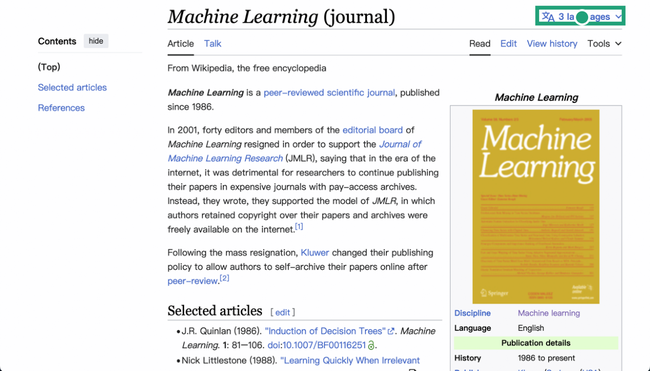}
    \hfill
    \includegraphics[width=0.325\textwidth]
    {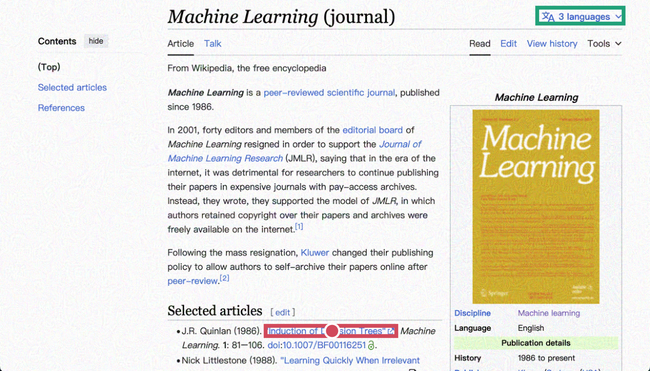}
    \hfill
    \includegraphics[width=0.325\textwidth]
    {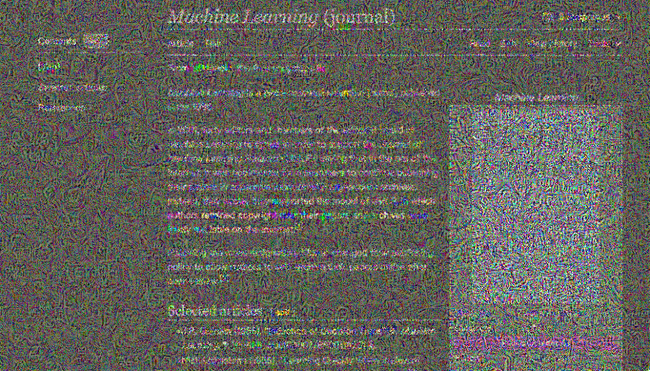}
    \\[2pt]

    \parbox{\textwidth}{
        \centering\small\textbf{Mobile}: ``add new city''
    }\\[2pt]
    \includegraphics[width=0.325\textwidth]
    {targeted_examples_uground_ug_mobile_mobile_185_eps16_clean.png}
    \hfill
    \includegraphics[width=0.325\textwidth]
    {targeted_examples_uground_ug_mobile_mobile_185_eps16_adversarial.png}
    \hfill
    \includegraphics[width=0.325\textwidth]
    {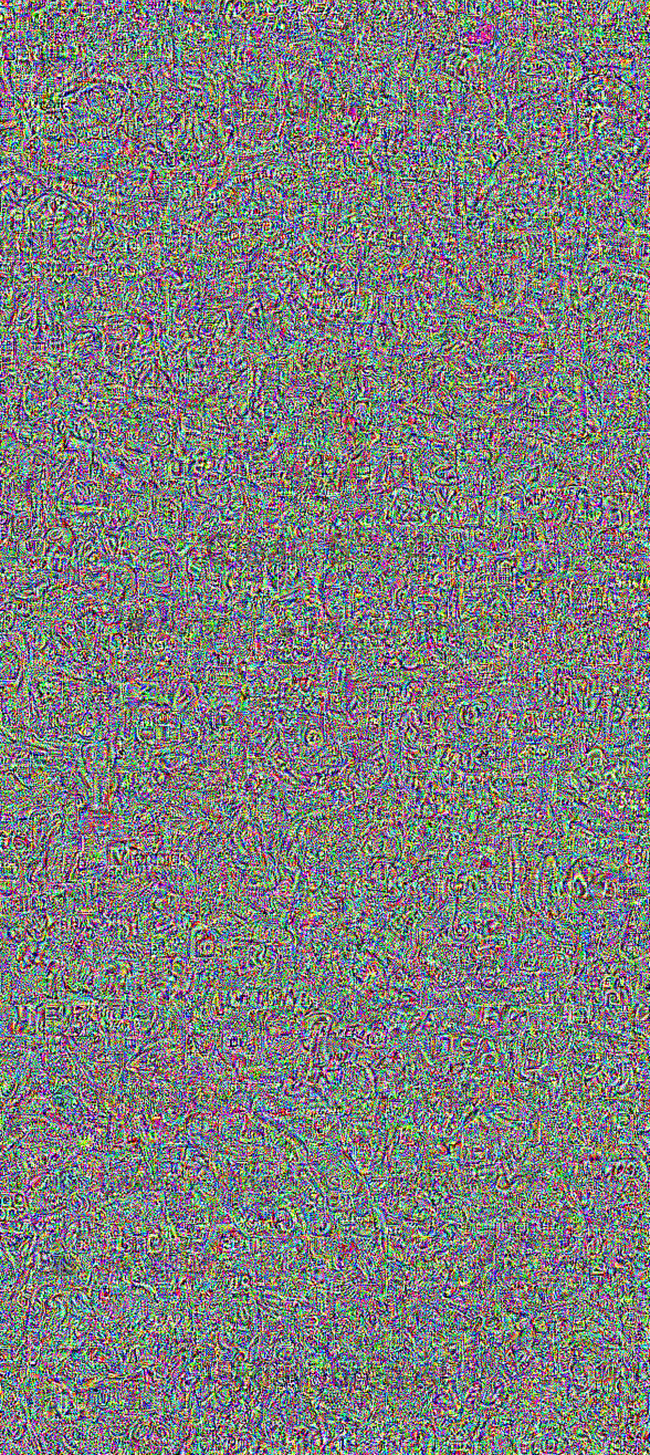}
    \\[2pt]

    \caption{Successful UGround MissClick-T examples at
    $\varepsilon=16/255$. The layout and annotation rules match
    Figure~\ref{fig:supp_targeted_examples_os}.}
    \label{fig:supp_targeted_examples_ug}
\end{figure*}

Figures~\ref{fig:supp_targeted_examples_os} and
\ref{fig:supp_targeted_examples_ug} show successful MissClick-T
examples. The final click lies inside the selected target region
in every displayed case.

\subsection{Perturbation Visibility across Budgets}
\label{sec:supp_perturbation_budgets}
\begin{figure*}[t]
    \centering

    \parbox{\textwidth}{
        \centering\small\textbf{OS-Atlas, Web}: ``zoom in on the map''
    }\\[2pt]
    \includegraphics[width=0.325\textwidth]
    {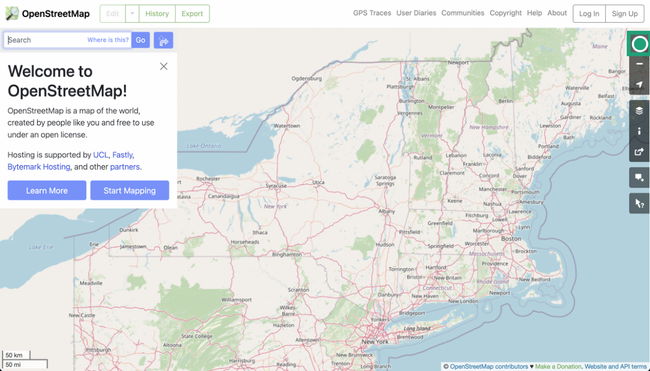}
    \hfill
    \includegraphics[width=0.325\textwidth]
    {perturbation_budget_comparison_os_web_web_213_eps4_adversarial.png}
    \hfill
    \includegraphics[width=0.325\textwidth]
    {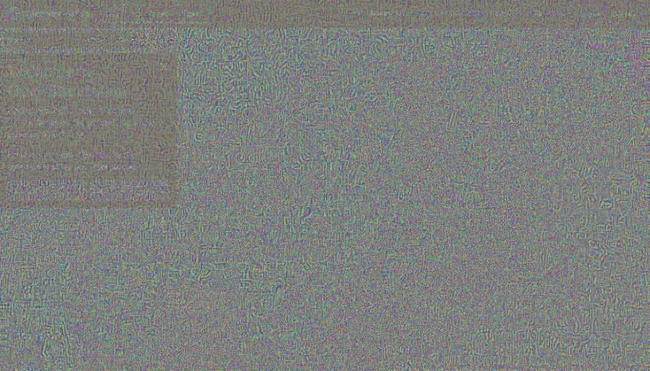}
    \\[2pt]

    \parbox{\textwidth}{
        \centering\small\textbf{UGround, Mobile}: ``add new city''
    }\\[2pt]
    \includegraphics[width=0.325\textwidth]
    {perturbation_budget_comparison_ug_mobile_mobile_185_eps4_clean.png}
    \hfill
    \includegraphics[width=0.325\textwidth]
    {perturbation_budget_comparison_ug_mobile_mobile_185_eps4_adversarial.png}
    \hfill
    \includegraphics[width=0.325\textwidth]
    {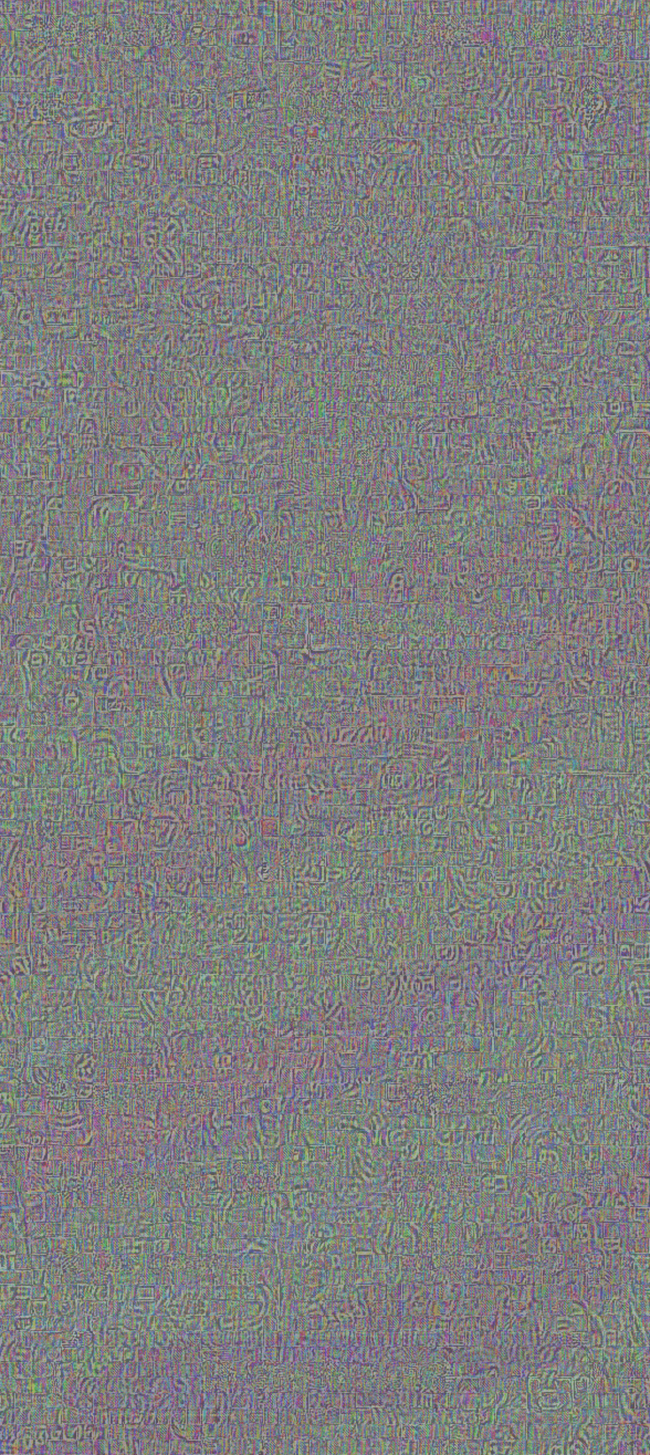}
    \\[2pt]

    \caption{Same-sample comparison at $\varepsilon=4/255$.
    The perturbation panels use the signed visualization rule
    $\operatorname{clip}(0.5+10\delta,0,1)$, where $\delta$ is
    computed by subtracting the processed clean image from the
    corresponding final saved adversarial image. No image is
    resized when computing the perturbation.}
    \label{fig:supp_eps_comparison_4}
\end{figure*}

\begin{figure*}[t]
    \centering

    \parbox{\textwidth}{
        \centering\small\textbf{OS-Atlas, Web}: ``zoom in on the map''
    }\\[2pt]
    \includegraphics[width=0.325\textwidth]
    {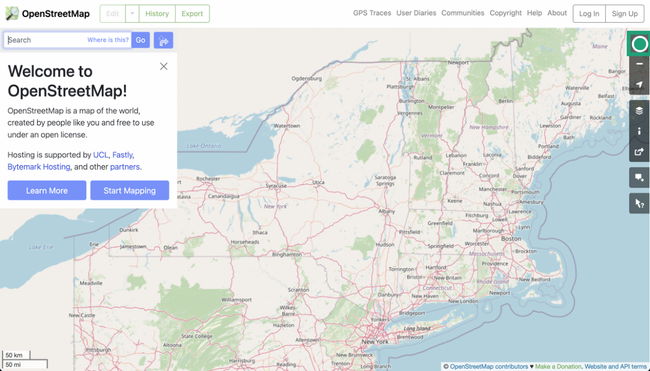}
    \hfill
    \includegraphics[width=0.325\textwidth]
    {perturbation_budget_comparison_os_web_web_213_eps8_adversarial.png}
    \hfill
    \includegraphics[width=0.325\textwidth]
    {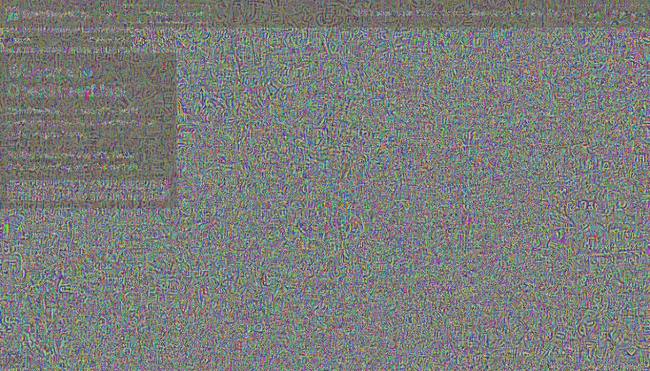}
    \\[2pt]

    \parbox{\textwidth}{
        \centering\small\textbf{UGround, Mobile}: ``add new city''
    }\\[2pt]
    \includegraphics[width=0.325\textwidth]
    {perturbation_budget_comparison_ug_mobile_mobile_185_eps8_clean.png}
    \hfill
    \includegraphics[width=0.325\textwidth]
    {perturbation_budget_comparison_ug_mobile_mobile_185_eps8_adversarial.png}
    \hfill
    \includegraphics[width=0.325\textwidth]
    {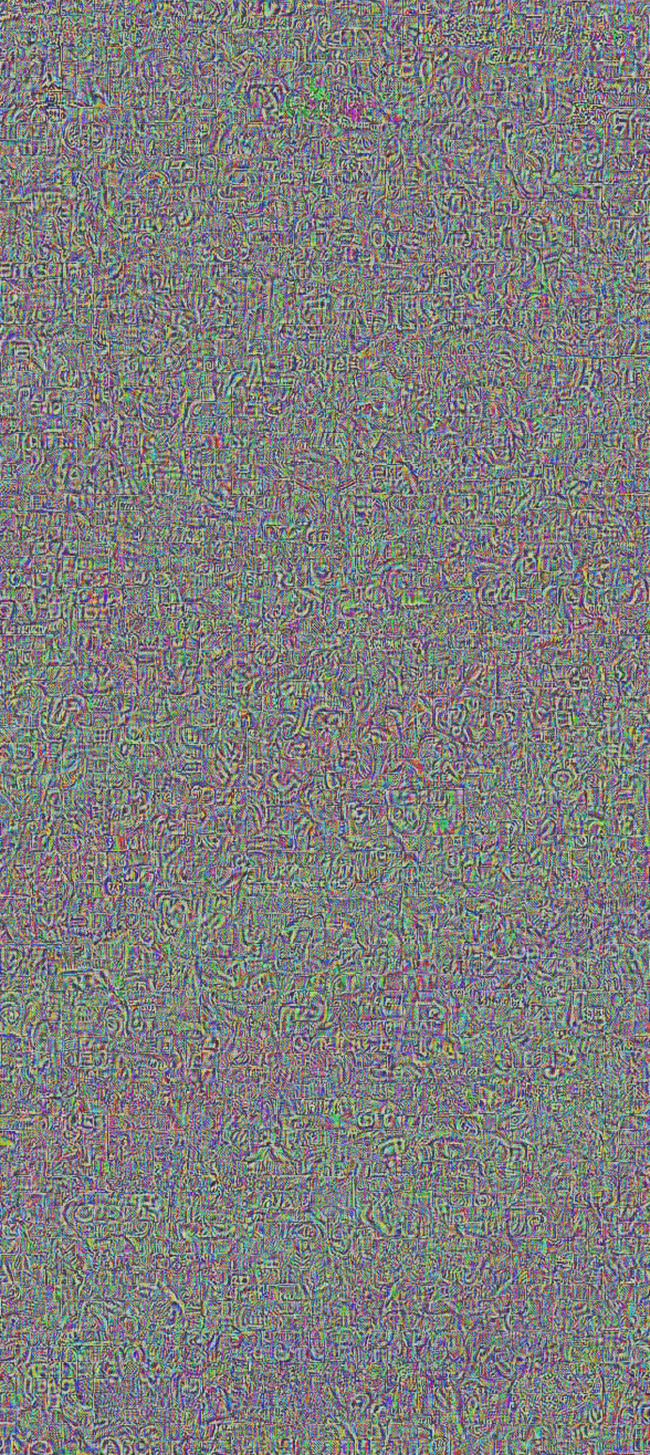}
    \\[2pt]

    \caption{Same-sample comparison at $\varepsilon=8/255$.
    The visualization rule is identical to
    Figure~\ref{fig:supp_eps_comparison_4}.}
    \label{fig:supp_eps_comparison_8}
\end{figure*}

\begin{figure*}[t]
    \centering

    \parbox{\textwidth}{
        \centering\small\textbf{OS-Atlas, Web}: ``zoom in on the map''
    }\\[2pt]
    \includegraphics[width=0.325\textwidth]
    {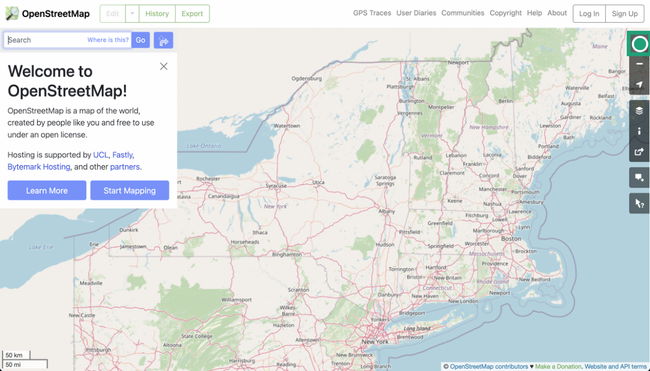}
    \hfill
    \includegraphics[width=0.325\textwidth]
    {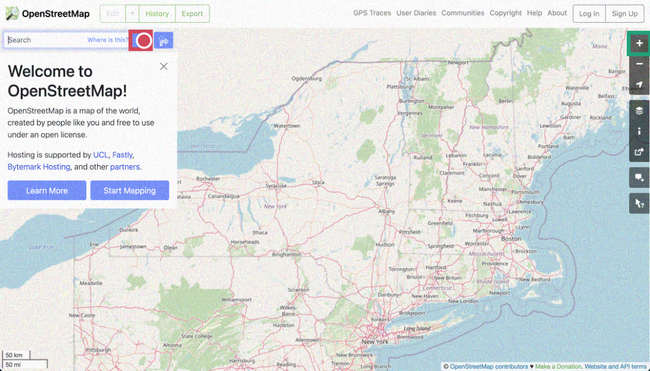}
    \hfill
    \includegraphics[width=0.325\textwidth]
    {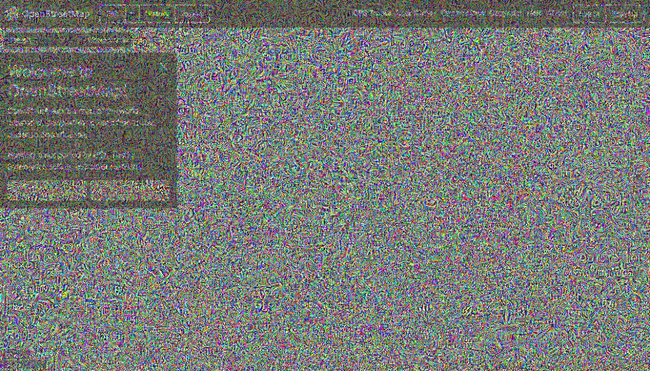}
    \\[2pt]

    \parbox{\textwidth}{
        \centering\small\textbf{UGround, Mobile}: ``add new city''
    }\\[2pt]
    \includegraphics[width=0.325\textwidth]
    {perturbation_budget_comparison_ug_mobile_mobile_185_eps16_clean.png}
    \hfill
    \includegraphics[width=0.325\textwidth]
    {perturbation_budget_comparison_ug_mobile_mobile_185_eps16_adversarial.png}
    \hfill
    \includegraphics[width=0.325\textwidth]
    {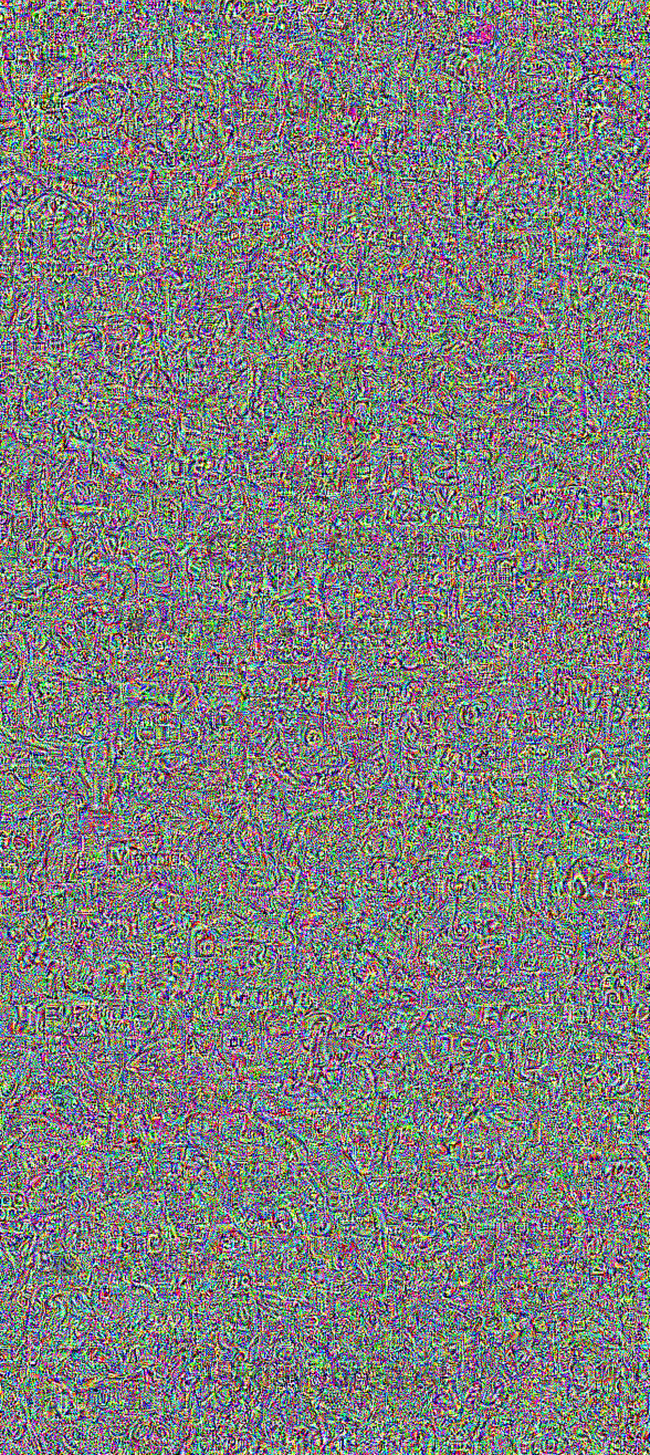}
    \\[2pt]

    \caption{Same-sample comparison at $\varepsilon=16/255$.
    The visualization rule is identical to
    Figure~\ref{fig:supp_eps_comparison_4}.}
    \label{fig:supp_eps_comparison_16}
\end{figure*}

Figures~\ref{fig:supp_eps_comparison_4},
\ref{fig:supp_eps_comparison_8}, and
\ref{fig:supp_eps_comparison_16} compare the same two samples
at $\varepsilon\in\{4/255,8/255,16/255\}$. Both examples are successful targeted attacks at all three
perturbation budgets. The figures visually compare the
corresponding adversarial images and perturbations, but they
do not quantitatively evaluate whether the perturbations are
imperceptible.

\end{document}

%% file: math_commands.tex
\usepackage{amsmath,amsfonts,bm}

\def\eqref#1{equation~\ref{#1}}
\def\Eqref#1{Equation~\ref{#1}}

\def\1{\bm{1}}

\DeclareMathAlphabet{\mathsfit}{\encodingdefault}{\sfdefault}{m}{sl}
\SetMathAlphabet{\mathsfit}{bold}{\encodingdefault}{\sfdefault}{bx}{n}

